\documentclass[runningheads]{llncs}

\usepackage{eccv}

\usepackage{eccvabbrv}

\usepackage{graphicx}
\usepackage{multirow} 
\usepackage{makecell} 
\usepackage{colortbl} 
\usepackage{diagbox} 
\usepackage{siunitx}
\usepackage{enumitem} 
\usepackage{arydshln} 
\usepackage{enumitem} 
\usepackage{arydshln} 
\usepackage{tabularx}
\usepackage{booktabs} 
\usepackage{xspace} 
\usepackage[Export]{adjustbox} 
\usepackage{pifont} 
\usepackage{dsfont}
\usepackage[accsupp]{axessibility}  
\usepackage{bbding}
\usepackage[dvipsnames]{xcolor} 

\usepackage{hyperref}

\def\eg{\emph{e.g.}\xspace} 
\def\ie{\emph{i.e.}\xspace}

\def\etal{\emph{et al.}\xspace}

\newcommand{\ours}{GlORIE-SLAM\xspace}
\newcommand{\boldparagraph}[1]{\vspace{0pt}\noindent{\bf #1}}

\newcommand{\greencheck}{{\color{OliveGreen}\checkmark}}
\newcommand{\redx}{{\color{red}\ding{55}}}

\colorlet{colorFst}{Green!25}       
\colorlet{colorSnd}{SpringGreen!45} 
\colorlet{colorTrd}{Yellow!30}      
\colorlet{colorLow}{darkgray!30}    
\newcommand{\fst}{\cellcolor{colorFst}\bf}   
\newcommand{\nd}{\cellcolor{colorSnd}}      
\newcommand{\rd}{\cellcolor{colorTrd}}      

\definecolor{gray}{rgb}{0.65,0.65,0.65}
\definecolor{mycol}{rgb}{0.90,0.95,1.0}

\usepackage{orcidlink}

\newcommand{\authorinfo}{
\author{Ganlin Zhang\inst{1*} \and
Erik Sandström\inst{1*} \and
Youmin Zhang\inst{2,3} \and 
Manthan Patel\inst{1}, \\ 
Luc Van Gool\inst{1,4,5} \and
Martin R. Oswald\inst{1,6}
}

\authorrunning{G.~Zhang$^{*}$, E.~Sandström$^{*}$, \etal}

\institute{$^1$\,ETH Zürich,  
$^2$\,University of Bologna, 
$^3$\,Rock Universe, 
$^4$\,KU Leuven, \\ 
$^5$\,INSAIT, 
$^6$\,University of Amsterdam
}
\renewcommand{\thefootnote}{\fnsymbol{footnote}}
\footnotetext[1]{The two authors contributed  equally to this paper.}
}

\begin{document}

\title{GlORIE-SLAM: Globally Optimized RGB-only Implicit Encoding Point Cloud SLAM}


\titlerunning{GlORIE-SLAM}

\authorinfo

\maketitle


{%
\footnotesize  
\begin{figure}
\vspace{-2em}
  \centering
  \resizebox{1.0\columnwidth}{!}{
  \newcommand{\sz}{0.3}
  \newcommand{\halfsz}{0.16}
  \renewcommand{\arraystretch}{1}
  \setlength{\tabcolsep}{1.0pt}
  \vspace{-2em}
  \small
  \begin{tabular}{cccc}
    \rotatebox{90}{\hspace{2em}\text{Textured Mesh}} \; &
    \includegraphics[width=\sz\linewidth]{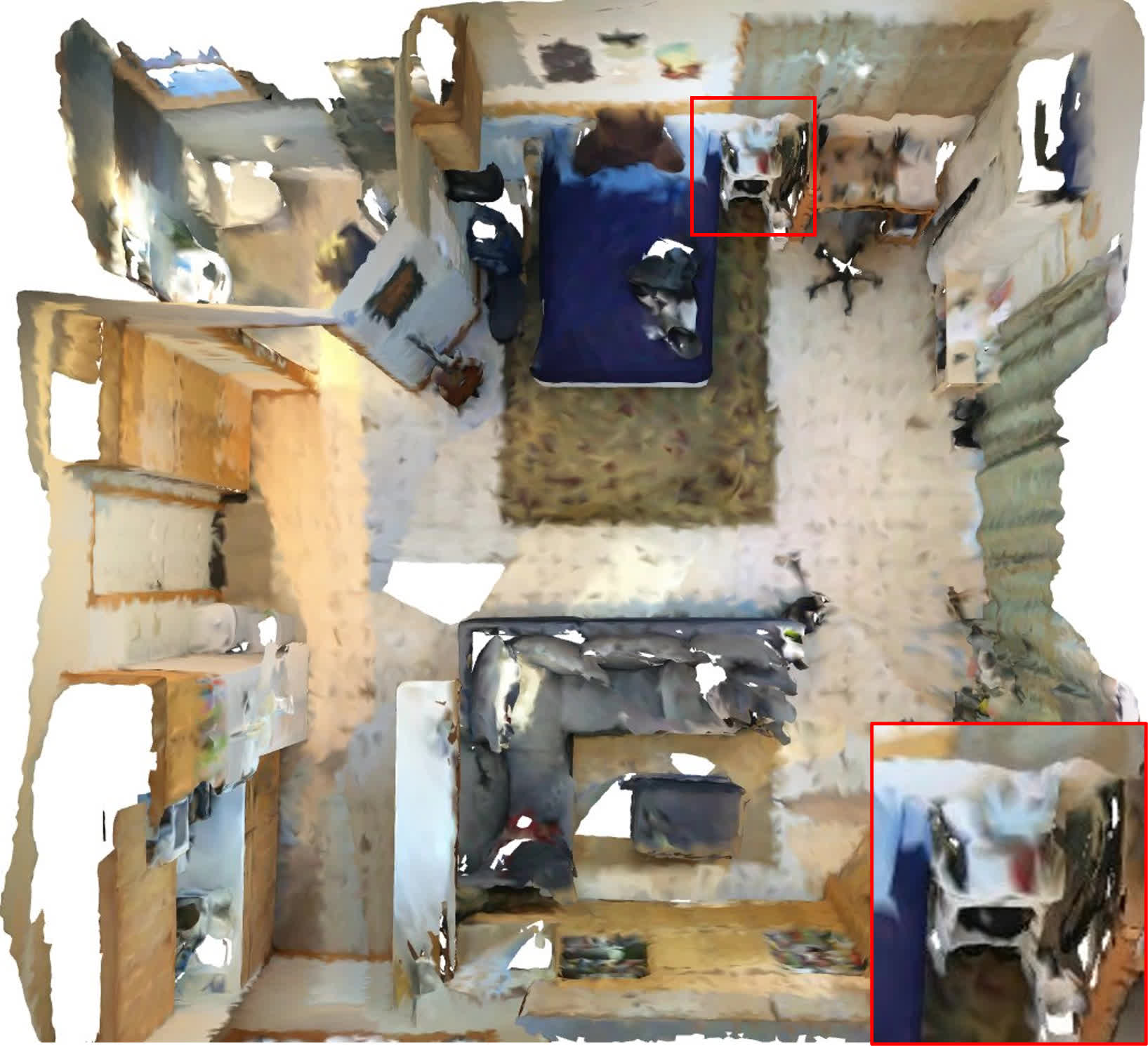} & 
    \includegraphics[width=\sz\linewidth]{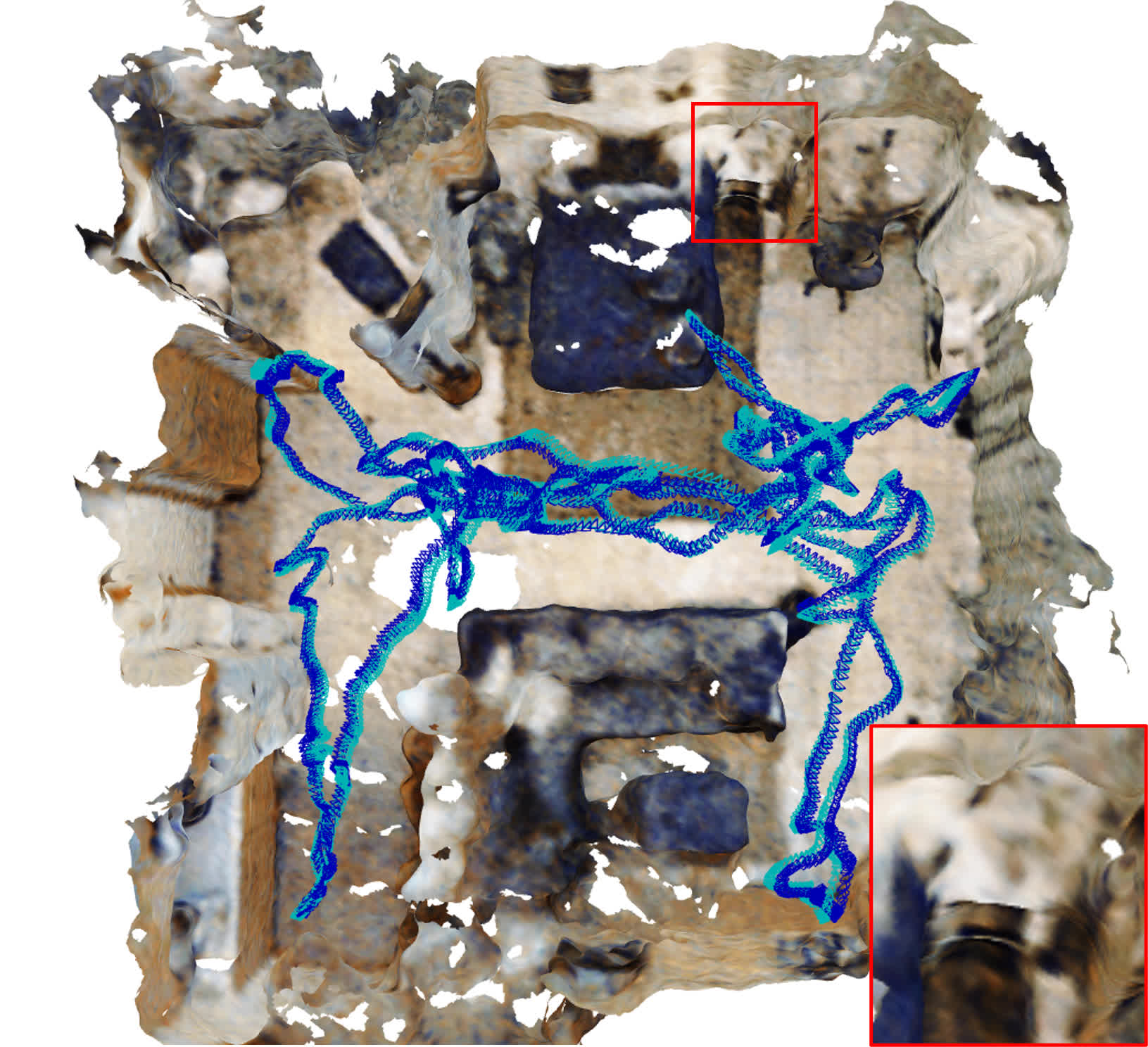} & 
    \includegraphics[width=\sz\linewidth]{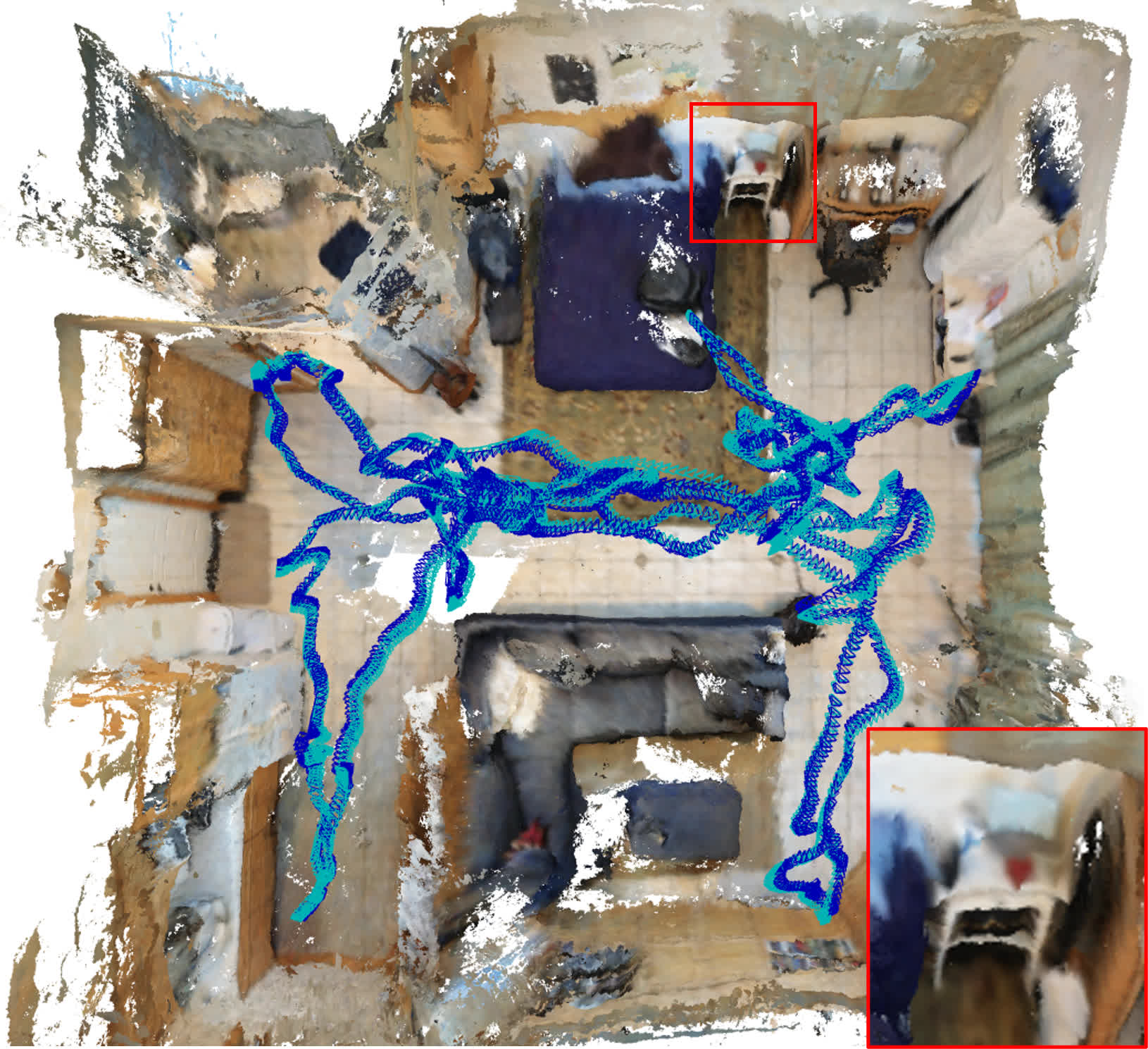}
    \\
    \rotatebox{90}{\hspace{0em}\text{Rendered}} \; &
    \includegraphics[width=\halfsz\linewidth]{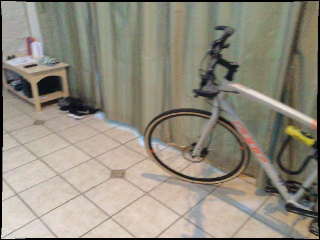} \hspace{-6.2pt}     \includegraphics[width=\halfsz\linewidth]{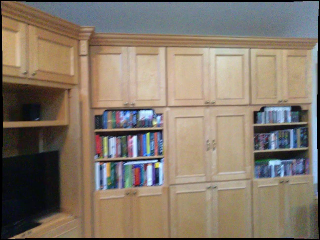}& 
    \includegraphics[width=\halfsz\linewidth]{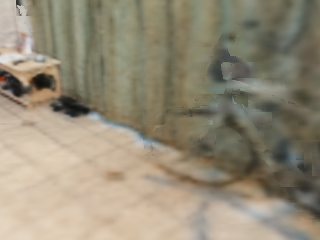} \hspace{-6.2pt}   \includegraphics[width=\halfsz\linewidth]{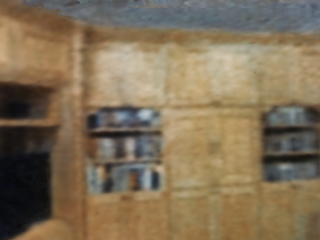} & 
    \includegraphics[width=\halfsz\linewidth]{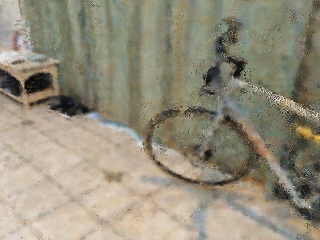} \hspace{-6.2pt} 
    \includegraphics[width=\halfsz\linewidth]{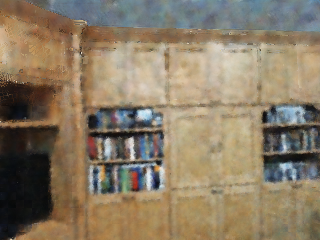}\\
    [0pt]
    &\small Ground Truth &
    \small GO-SLAM~\cite{zhang2023go} & 
    \small \textbf{\ours (Ours)}\\
    [0pt]
    &\makecell[c]{\small ATE RMSE / PSNR:}
    &
    \small 5.9 \texttt{cm} / 15.74 \texttt{dB} & 
    \small \textbf{5.5 \texttt{cm} / 23.42 \texttt{dB}}\\
    
  \end{tabular}}
  \caption{\textbf{\ours Results on ScanNet scene \texttt{0000}.} \ours uses a deformable point cloud as the scene representation and 
  achieves lower trajectory error and higher rendering accuracy compared to competitive approaches, \eg GO-SLAM. The geometric accuracy is qualitatively evaluated. The \textcolor{Cerulean}{light blue} trajectory is ground truth and the \textcolor{blue}{blue} is the estimated. The PSNR is evaluated for all keyframes.}
  \label{fig:teaser}
\end{figure}
\vspace{-3em}
}
\begin{abstract}
  Recent advancements in RGB-only dense Simultaneous Localization and Mapping (SLAM) have predominantly utilized grid-based neural implicit encodings and/or struggle to efficiently realize global map and pose consistency. To this end, we propose an efficient RGB-only dense SLAM system using a flexible neural point cloud scene representation that adapts to keyframe poses and depth updates, without needing costly backpropagation. 
  Another critical challenge of RGB-only SLAM is the lack of geometric priors. To alleviate this issue, with the aid of a monocular depth estimator, we introduce a novel DSPO layer for bundle adjustment which optimizes the pose and depth of keyframes along with the scale of the monocular depth.
  Finally, our system benefits from loop closure and online global bundle adjustment and performs either better or competitive to existing dense neural RGB SLAM methods in tracking, mapping and rendering accuracy on the Replica, TUM-RGBD and ScanNet datasets. The source code is available at \url{https://github.com/zhangganlin/GlOIRE-SLAM}.
\end{abstract}

\section{Introduction}
The past few years have witnessed a tremendous growth of dense SLAM, fueled predominantly by ideas such as extrapolation capability in occluded regions~\cite{Sucar2021IMAP:Real-Time,zhu2022nice}, joint geometric and semantic 3D prediction~\cite{zhu2023sni,li2023dns}, using learnable priors for increased tracking and mapping robustness \cite{zhu2023nicer,zhang2023hi,naumann2023nerf,zhou2024modslam}, improving compression of the 3D scene representation~\cite{Sucar2021IMAP:Real-Time,zhu2022nice,yang2022vox,mahdi2022eslam,sandstrom2023point} and online uncertainty estimation \cite{uncleslam2023,Rosinol2022NeRF-SLAM:Fields}. A common factor for the majority of these works is that they reconstruct the dense map by optimizing a neural implicit encoding of the scene in the form of either weights of an MLP~\cite{azinovic2022neural,Sucar2021IMAP:Real-Time,matsuki2023imode,ortiz2022isdf} or as features anchored in dense grids~\cite{zhu2022nice,newcombe2011kinectfusion,Weder2020RoutedFusion,weder2021neuralfusion,sun2021neuralrecon,bovzivc2021transformerfusion,li2022bnv,zou2022mononeuralfusion,uncleslam2023}, hierarchical octrees~\cite{yang2022vox}, via voxel hashing~\cite{zhang2023go,zhang2023hi,chung2022orbeez,Rosinol2022NeRF-SLAM:Fields,matsuki2023newton}, point clouds \cite{hu2023cp,sandstrom2023point,liso2024loopyslam} or axis aligned feature planes~\cite{mahdi2022eslam,peng2020convolutional}. Concurrent to our work, we have also seen the introduction of 3D Gaussian Splatting (3DGS) to the dense SLAM field~\cite{yugay2023gaussianslam,keetha2023splatam,yan2023gs,matsuki2023gaussian,huang2023photo}.

When surveying the recent trends, we make certain discoveries: 
\textbf{1.} Explicit representations such as point clouds achieve state-of-the-art rendering and reconstruction accuracy~\cite{sandstrom2023point}.
\textbf{2.} The majority of works perform frame-to-model tracking and struggle to efficiently introduce bundle adjustment (BA) and/or loop closure~\cite{sandstrom2023point,mahdi2022eslam,Wang_2023_CVPR,tofslam,Hu2023LNI-ADFP,li2023end,zhu2023nicer,uncleslam2023,li2023dense}. Instead, the methods with the lowest camera pose drift on real-world data are based on external trackers such as DROID-SLAM~\cite{teed2021droid}, extended with online loop closure and global BA~\cite{zhang2023go,zhang2023hi}, or ORB-SLAM2~\cite{chung2022orbeez}. 
\textbf{3.} The RGB-only setting is under-explored compared to the RGB-D setting as it needs to deal with depth and scale ambiguities, making the problem more difficult. Recently, HI-SLAM~\cite{zhang2023hi} and GO-SLAM~\cite{zhang2023go} introduce dense RGB-only SLAM frameworks. They use hierarchical feature grids as the dense map representation, which does not efficiently and accurately lend itself for map deformations. 
Such map deformations are required when performing loop closure or BA to avoid recomputing the map from scratch and hence requiring to save all input data as commonly done in the literature~\cite{innmann2016volumedeform,Rosinol2022NeRF-SLAM:Fields}.

To this end, we propose to use an efficient and accurate neural point cloud scene representation to construct an RGB-only SLAM system that can also handle complex and large scale indoor scenes by integrating online loop closure along with online global bundle adjustment (BA) and a monocular depth prior which aids not only in reconstruction completeness, but also in accuracy.



\noindent In summary, our \textbf{contributions} include:
\begin{itemize}[itemsep=0pt,topsep=2pt,leftmargin=10pt,label=$\bullet$] 
  \item An RGB-only dense SLAM framework which performs camera pose estimation via dense optical flow tracking and includes loop closure and online global BA. Mapping is done with a neural point cloud map representation that allows for online rigid map deformations (instead of costly map re-creation) for online loop closures as well as pose and geometry refinement. 
  This results in more accurate reconstructions compared to hierarchical grid-based approaches while not requiring backpropagation to retrain the grid-anchored features. See \cref{fig:teaser}.
  \item A novel Disparity, Scale and Pose Optimization (DSPO) layer that combines pose and geometry estimation. It refines inaccurate parts of the estimated keyframe disparity by tightly coupling a monocular depth prior into the bundle adjustment.
\end{itemize}
\section{Related Work} \label{sec:rel}

\boldparagraph{Dense Visual SLAM.} 
The pioneering work of Curless and Levoy~\cite{curless1996volumetric} using truncated signed distance functions (TSDF) laid the foundation for dense online 3D reconstruction strategies. KinectFusion~\cite{newcombe2011kinectfusion} demonstrated real-time dense mapping and tracking using depth maps and scalability was further improved through techniques like voxel hashing~\cite{niessner2013voxel_hashing,Kahler2015infiniTAM,Oleynikova2017voxblox,dai2017bundlefusion,matsuki2023newton},
and octrees~\cite{steinbrucker2013large,yang2022vox,marniok2017efficient,chen2013scalable,liu2020neural}. Besides grid based methods, point based SLAM has also been successful~\cite{whelan2015elasticfusion,schops2019bad,cao2018real,Kahler2015infiniTAM,keller2013real,cho2021sp,zhang2020dense,sandstrom2023point}, typically in the form of surface elements (surfels).
To obtain globally consistent pose estimates and dense maps, numerous techniques have been developed. The most popular includes splitting the global map into parts, usually called submaps~\cite{cao2018real,dai2017bundlefusion,fioraio2015large,tang2023mips,matsuki2023newton,maier2017efficient,kahler2016real,stuckler2014multi,choi2015robust,Kahler2015infiniTAM,reijgwart2019voxgraph,henry2013patch,bosse2003atlas,maier2014submap,tang2023mips,mao2023ngel}, which aggregate the information from a set of frames. Once a loop is detected, the submaps are rigidly registered via pose graph optimization~\cite{cao2018real,maier2017efficient,tang2023mips,matsuki2023newton,kahler2016real,endres2012evaluation,engel2014lsd,kerl2013dense,choi2015robust,henry2012rgb,yan2017dense,schops2019bad,reijgwart2019voxgraph,henry2013patch,stuckler2014multi,wang2016online,matsuki2023newton,hu2023cp,mao2023ngel}. To refine the solution, global BA is sometimes applied~\cite{dai2017bundlefusion,schops2019bad,cao2018real,teed2021droid,yan2017dense,yang2022fd,matsuki2023newton,chung2022orbeez,tang2023mips,hu2023cp}. 

Most similar to our work, but in the RGBD setting, is the concurrent Loopy-SLAM~\cite{liso2024loopyslam}, which also utilizes the map representation from Point-SLAM~\cite{sandstrom2023point}. They split the global map into submaps
and apply a robust pose graph formulation based on direct point cloud registration. Other concurrent RGBD works include methods that use 3D Gaussians as the scene representation~\cite{yugay2023gaussianslam,keetha2023splatam,yan2023gs}. None of these works consider global map consistency via \eg loop closure. For a recent in depth survey on NeRF inspired dense SLAM, we refer to~\cite{tosi2024nerfs}.

\boldparagraph{RGB-only Dense Visual SLAM.}
The majority of research focuses on combining RGB and depth cameras, while methods using only RGB cameras are less explored due to their lack of geometric data, causing scale and depth ambiguities. However, RGB-only dense SLAM is appealing for its cost-effectiveness and versatility in various environments.
Within the scope of NeRF inspired dense SLAM, NeRF-SLAM~\cite{Rosinol2022NeRF-SLAM:Fields} and Orbeez-SLAM~\cite{chung2022orbeez} use DROID-SLAM~\cite{teed2021droid} and ORB-SLAM2~\cite{mur2017orb} as the tracking modules respectively. Both use Instant-NGP~\cite{muller2022instant} as the volumetric neural radiance field map. NeRF-SLAM does not tackle the global consistency of the map and Orbeez-SLAM~\cite{chung2022orbeez} requires undesirable training iterations of the hash features to perform map corrections. DIM-SLAM~\cite{li2023dense} and NICER-SLAM~\cite{zhu2023nicer} both perform tracking and mapping on the same neural implicit map representation, consisting of hierarchical feature grids and do not address global map consistency via \eg loop closure. Recently, GO-SLAM~\cite{zhang2023go} and Hi-SLAM~\cite{zhang2023hi} adapts DROID-SLAM~\cite{teed2021droid} to the full SLAM setting by introducing online loop closure via factor graph optimization. Both systems achieve high tracking accuracy, but like Orbeez-SLAM and NeRF-SLAM, they require backpropagation steps to update the feature encodings in the Instant-NGP map. We are inspired by both works, and make some important modifications to increase tracking and mapping accuracy. We integrate our own point based map representation to account for online updates of poses and geometry as a result of bundle adjustment (BA) or loop closure.

Concurrent to our work, FMapping~\cite{hua2023fmapping}, Hi-Map~\cite{hua2024hi}, TT-HO-SLAM~\cite{naumann2023nerf} and NeRF-VO~\cite{naumann2023nerf} all utilize hierarchical grid based data structures. MoD-SLAM uses an MLP to parameterize the map via a unique reparameterization. 
Perhaps most closely related to our work are MonoGS~\cite{matsuki2023gaussian} and Photo-SLAM \cite{huang2023photo}. They both tackle monocular dense SLAM by modeling the map with 3D Gaussians.

\boldparagraph{Depth Priors for RGB-only SLAM.} NICER-SLAM~\cite{zhu2023nicer} estimates the scale and shift of a relative mono-depth estimator and supervises all pixels equally. MoD-SLAM \cite{zhou2024modslam} combines relative and metric mono-depth estimation, and requires additional finetuning of the metric depth. HI-SLAM~\cite{zhang2023hi} proposes a similar technique to ours, but regularizes all available keyframe depth pixels with the mono-depth prior. In our DSPO layer, we instead split the optimization and use the monocular prior to regularize the high error keyframe depth pixels while the low error keyframe depth is kept fixed to stabilize scale estimation.



\begin{figure}[t!]
\centering
 \includegraphics[width=1.0\linewidth]{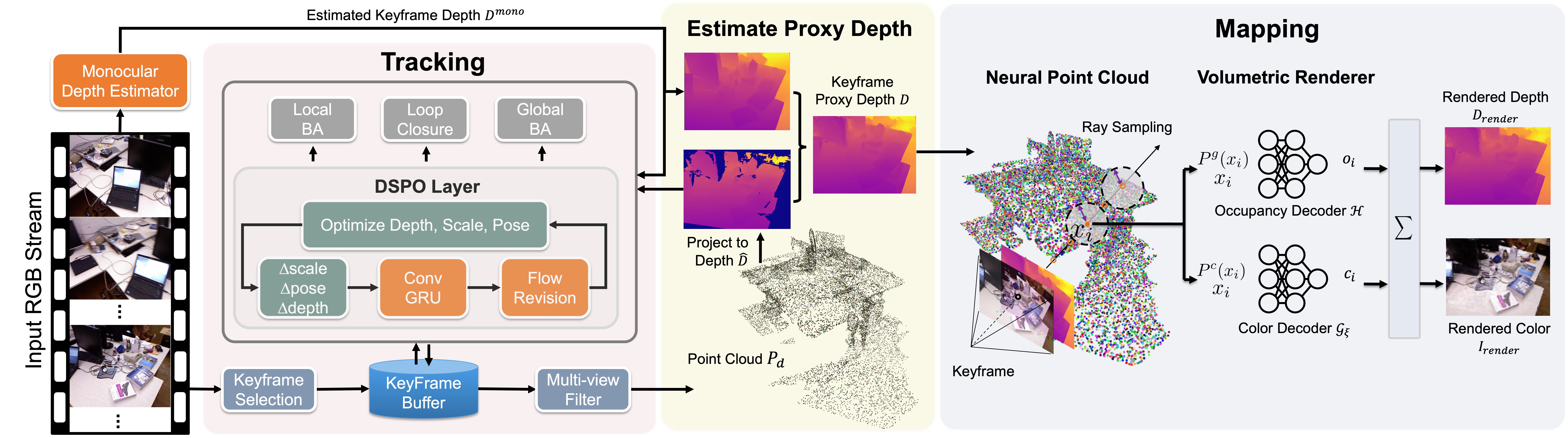}\\
\caption{\textbf{\ours{} Architecture.} Given an input RGB stream, we first track and then map every keyframe. The pose is initially estimated with local bundle adjustment (BA) via frame-to-frame tracking of recurrent optical flow estimation. This is done with our novel DSPO (Disparity, Scale and Pose Optimization) layer, which combines pose and depth estimation with scale and depth refinement by leveraging a monocular depth prior. The DSPO layer also refines the poses globally via online loop closure and global BA. To map the estimated pose, a proxy depth map is estimated by combining the noisy keyframe depths from the tracking module with the monocular depth prior to account for missing observations. Mapping is done, along with the input RGB keyframe via a deformable neural point cloud, leveraging depth guided volumetric rendering. A re-rendering loss to the input RGB and proxy depth optimizes the neural features and the color decoder weights. Importantly, the neural point cloud deforms to account for global updates of the poses and proxy depth before each mapping phase.
}
\label{fig:architecture}
\end{figure}
\section{Method} \label{sec:methods}

As an RGB-only SLAM framework, our aim is to simultaneously track the camera poses and reconstruct globally consistent scene geometry. To achieve this, we utilize a deformable neural point cloud as the scene representation (\cref{sec:npc_rep}). During mapping, we render depth and color from the neural point cloud (\cref{sec:rendering}), and optimize a re-rendering loss with respect to the sensor input (\cref{sec:mapping_loss}). For tracking, we use an optical-flow-based method containing loop closure and online global bundle adjustment. The tracker takes an RGB stream along with relative depth from a monocular depth estimator as input and outputs is a set of estimated camera poses and noisy semi-dense depth maps and feeds them to the mapper (\cref{sec:tracking}). For an overview of our method, see \cref{fig:architecture}.

\subsection{Neural Point Cloud Scene Representation} \label{sec:npc_rep}
Similar to Point-SLAM~\cite{sandstrom2023point}, we use a neural point cloud to represent the scene, but the RGB-only setting requires several modifications.
The neural point cloud with $N$ neural points is defined as 
\begin{equation}
  P = \{\left(p_i, f^g_i, f^c_i, k_i, u_i, v_i, D_i \right) \, | \, i=1,\ldots,N\} \enspace,
\end{equation}
with point locations $p_i \in \mathbb{R}^3$ and $f^g_i, f^c_i \in \mathbb{R}^{32}$ being the geometric and color features encoding the local shape and appearance information of $p_i$ and its surrounding.
The other attributes are used to deform the map at \eg loop closure. $k_i$ is the frame index that anchored point $i$ from pixel ($u_i$, $v_i$) at depth $D_i$.

\boldparagraph{Point Adding Strategy.} 
For every keyframe, given the estimated camera pose and proxy depth map (see \cref{sec:rendering}), we adopt the point adding strategy from Point-SLAM \ie we sample $X$ pixels uniformly and $Y$ pixels among the top $5Y$ pixels according to the color gradient magnitude. This ensures an even spread of pixels while focusing on reconstructing complex areas properly. The $X+Y$ pixels are projected into 3D space, and if no existing point is found within radius $r$, three neural points are initialized along the ray, with depth $(1-\rho)D$, $D$ and $(1+\rho)D$ respectively, with $D$ being the proxy depth of the pixel and $\rho \in (0,1)$ is a hyperparameter. 
The search radius $r$ is computed dynamically as a linear transformation of the pixel color gradient $\nabla I(u,v)$, and clamped by the predefined maximum and minimum radius $r_u$ and $r_l$ respectively. Different from Point-SLAM, the scale of the scene cannot be determined a priori in the RGB-only setting. 
We therefore scale the radius with the estimated depth to adapt to the reconstructed scene size as
\begin{equation}
  r(u,v)= D(u,v)\max\big(\min(\beta_1\nabla I(u,v) + \beta_2,\, r_u),\, r_l\big)  \enspace,
  \label{eq:r}
\end{equation}
where $\beta_1, \beta_2 \in\mathbb{R}$ are hyperparameters.

\boldparagraph{Map Update.} 
In a SLAM system, the estimated set of camera poses are globally updated as a result of bundle adjustment and loop closure. For this reason, the map needs to be deformed to ensure it is consistent with the updated poses. Thanks to our point-based representation, map deformations are straightforward. For every tuple of three points, which are added along a ray in the neural point cloud, we update the locations $p_i$ by rigidly re-anchoring them. Specifically, for the central point, the new point location ${p_i}'$ is computed as,
\begin{equation}
    p^\prime_i = \omega^\prime_{k_i} D^\prime_i K^{-1}[u_i, v_i, 1]^T,
\end{equation}
where ${\omega^\prime_{k_i}} \in \mathbb{SE}(3)$ is the updated camera-to-world pose, $D^\prime_i$ the updated depth and $K$ the temporally constant camera intrinsics. Note that only the locations of the points are updated and not the neural features.
For points that need to be deformed, but lack updated depth $D^\prime_i$, we rescale the depth map that first initialized and anchored the neural point. We do this with a least squares fitting to the new depth map such that $D^\prime_i = sD_i$, where $s$ is the rescaling factor and $D_i$ is the depth that was available when the point was initialized.

\subsection{Rendering} \label{sec:rendering}
Similar to Point-SLAM~\cite{sandstrom2023point}, we use depth-guided volume rendering that sparsely samples the 3D space for faster RGB and depth rendering.

\boldparagraph{Proxy Depth Map.}
Different from Point-SLAM, which benefits from a dense depth sensor, the estimated monocular depth is less reliable. 
To still benefit from sparse volume rendering, we construct a proxy depth map for each keyframe. We utilize the estimated dense but noisy depth map $\Tilde{D}$ from the tracker and further filter out inconsistent depth with multi-view constraints as follows.
For a given depth $\Tilde{D}_c$ from keyframe $c$, we project every pixel $(u,v)$ to 3D as $p_{c}$, which is then projected into keyframe $j$ to obtain the corresponding pixel $(\hat{u},\hat{v})$ as
\begin{equation}
        p_{c} = \omega_c\Tilde{D}_c(u, v)K^{-1}[u, v, 1]^T  ,\qquad  [\hat{u},\hat{v}, 1]^T \propto K\omega_j^{-1}[p_{c},1]^T \enspace.
\end{equation}
We unproject the depth at $(\hat{u},\hat{v})$ to obtain the corresponding 3D location $p_{j}$,
%
\begin{equation}
        p_{j} = \omega_j\Tilde{D}_j(\hat{u}, \hat{v})K^{-1}[\hat{u}, \hat{v}, 1]^T \enspace.
        \label{eq:back-project}
\end{equation}
If the L2 distance between $p_{c}$ and $p_{j}$ is small enough, the estimated depth $\Tilde{D}_c(u, v)$ is consistent between these two views. The global two-view consistency $n_{c}$ for frame $c$ can be computed by looping over all keyframes as
\begin{equation}
    n_{c}(u,v) = \sum_{\substack{k\in \text{KFs},\\ k\neq c}}
    \mathds{1}\Big(
    \left\| p_{c} - p_{k} \right\|
    < \eta\cdot\text{average}(\Tilde{D}_c)\Big) \enspace,
    \label{eq:two_view_consist}
\end{equation}
where $\mathds{1}(\cdot)$ is the indicator function. Here, $\eta \in\mathbb{R}_{\geq 0}$ is a hyperparameter and $n_{c}$ is the total two-view consistency for pixel $(u,v)$ in keyframe $c$. If $n_{c}$ is larger than a threshold, $\Tilde{D}_c(u, v)$ is valid.
%
We form a point cloud $P_d$ by unprojecting all depth values from the keyframes to 3D.
%
\begin{equation}
    P_d = \left\{ \omega_k\Tilde{D}_k(u, v)K^{-1}[u, v, 1]^T \ \big| \ k\in \text{KFs},\,\Tilde{D}_k(u,v)\, \text{ is valid} \right\}\enspace.
\end{equation}
Next, we utilize the filtered multi-view information in $P_d$ and form almost dense depth maps $\hat{D}_c$ by projecting $P_d$ into the keyframes. The more frames that are observed, the denser they become. Nevertheless, $\hat{D}_c$ can still have missing pixels. To fill these, we use a monocular depth prior $D^{\text{mono}}_c$ predicted by an off-the-shelf mono-depth estimator~\cite{eftekhar2021omnidata}. The predicted depth is, however, only estimated up to unknown scale $\theta_c$ and shift $\gamma_c$. These parameters are estimated by fitting the monocular depth to the depth maps $\hat{D}_c$ with least squares as
\begin{equation}
    \hat{\theta}_c,\,\hat{\gamma}_c = \mathop{\arg \min}_{\theta,\gamma}\sum_{(u,v)}\left((\theta D^{\text{mono}}_c(u,v)+\gamma) -\hat{D}_c(u,v) \right)^2 \enspace.
    \label{eq:wq_least_square}
\end{equation}
Then, the final proxy depth map $D_c$ is, 
\begin{equation}
  \!\!D_c(u,v)= 
  \begin{cases}
    \hat{D}_c(u,v) & \text{if $\hat{D}_c(u,v)$ has value}\\
    \hat{\theta}_c D^{\text{mono}}_c(u,v)+\hat{\gamma}_c & \text{otherwise .}\\
  \end{cases}
\end{equation}

\boldparagraph{Rendering RGB and Depth.} 
Given the estimated camera pose and corresponding proxy depth map, we sample a set of 3D points along a ray,
\begin{equation}
    p_i = o_c + z_i v_{i}\enspace,
\end{equation}
where $o_c \in\mathbb{R}^3$ is the camera origin, $z_i \in\mathbb{R}$ the point depth and $v_{i} \in\mathbb{R}^3$ the ray direction. We sample 10 points spread evenly between $(1-\rho)D_c$ and $(1+\rho)D_c$ ($D_c$ is the proxy depth) to ensure the real surface is contained within the extraction band, as the proxy depth can be noisy. 
For each sampled point $p_i$, two MLPs $\mathcal{H}$ and $\mathcal{G}_\xi$ decode the occupancy $s_i \in [0,1]$ and color $t_i \in\mathbb{R}^3$ respectively as
\begin{align} 
    {s}_{i} = \mathcal{H}\big(p_i, P^g(p_i)\big)\enspace, \qquad 
    t_i     = \mathcal{G}_\xi\big(p_i, P^c(p_i), v_i\big) \enspace.
\end{align}
For the occupancy decoder $\mathcal{H}$, we use the same MLP as Point-SLAM~\cite{sandstrom2023point} and use their pretrained and fixed parameters. For the color decoder $\mathcal{G}_\xi$, we modify the MLP used by~\cite{sandstrom2023point} by additionally feeding the view direction $v_i$ as input. 
The model parameters $\xi$ are optimized on the fly. To obtain the geometric feature $P^g$ and color feature $P^c$ at the sampled points $p_i$, we search neighboring neural points within the query radius $2r$ in the neural point cloud and use a distance-based weighted average of the features $f^g$ and $f^c$ of these neural points. We refer to~\cite{sandstrom2023point} for the details of the point cloud feature interpolation. The per-point occupancy and color are used to render the per-pixel depth and color via volume rendering. Specifically, we render the depth $D_{\text{render}}$ and color $I_{\text{render}}$ as the weighted average of the depth and color values along each ray,
\begin{align}
   D_{\text{render}} = \sum_{i=1}^N \alpha_i z_i  \enspace, \quad 
   I_{\text{render}} = \sum_{i=1}^N \alpha_i t_i  \enspace,
   \label{eq:rgbd-render}
\end{align}
where $\alpha_i = s_i\prod_{j=i}^{i-1}(1-s_j)$ can be interpreted as the probability that the ray terminates at point $p_i$, and $N$ is the number of sampled points along each ray.


\subsection{Loss Function} \label{sec:mapping_loss}
For each mapping phase, we first randomly select $\kappa$ keyframes which overlap  significantly with the viewing frustum of the current frame. 
Among the selected keyframes including the current frame, we sample $M$ pixels uniformly across the image plane similar to \cite{sandstrom2023point}. 
Rendering is done as described in \cref{sec:rendering} and we calculate the $L_1$ loss using the sensor RGB image and the proxy depth map,
\begin{align}
    \mathcal{L}_{\text{geo}}   = \sum_{m = 1}^M \lvert D_m - D^{\text{render}}_m \rvert_1 \enspace, \qquad 
    \mathcal{L}_{\text{color}} = \sum_{m = 1}^M \lvert I_m - I^{\text{render}}_m \rvert_1 \enspace.
\end{align}
We add an additional frame-to-frame pixel warping term to ensure geometric consistency of the map from different views. For each sampled pixel $(u,v)$ in keyframe $c$, we first project it to 3D as $p_m$, using the rendered depth $D_{\text{render}}$ as in \cref{eq:back-project}. Then $p_m$ is projected into other selected keyframes and the photometric loss between the projected pixel $(\hat{u},\hat{v})$ and the original pixel is applied:
\begin{equation}
    \mathcal{L}_{\text{pix}} = \sum_{m = 1}^M \sum_{\substack{k\in \text{KFs},\\ k\neq c}} \lvert I_c(u,v) - I_k(\hat{u},\hat{v}) \rvert_1 \enspace.
    \label{eq:pix_warp_loss}
\end{equation}
The final loss is
\begin{equation}
    \mathcal{L}= \lambda_{\text{geo}}\mathcal{L}_{\text{geo}} +  \lambda_{\text{pix}}\mathcal{L}_{\text{pix}} +  \lambda_{\text{color}}\mathcal{L}_{\text{color}} \enspace,
\end{equation}
with hyperparameters $\lambda_{\text{geo}}, \lambda_{\text{pix}},\lambda_{\text{color}} \in\mathbb{R}_{\geq 0}$. 
Importantly, we split the optimization into two stages. For the first 30 $\%$ of iterations, $\lambda_{color} = 0$ to initialize the geometry well. 
Then, $\lambda_{color} > 0$ and $f^g$, $f^c$ and $\xi$ are jointly optimized.


\subsection{Tracking} \label{sec:tracking}

To robustly track the camera under challenging scenarios, \eg repeated patterns, low texture or rapid camera movement, we adopt the idea of frame-to-frame tracking~\cite{zhang2023go} which is based on recurrent optical flow~\cite{teed2020raft}. 
A factor graph $\mathcal{G}(\mathcal{V},\mathcal{E})$ is maintained during the tracking process, where every node in $\mathcal{V}$ stores the pose and estimated disparity of a keyframe and every edge in $\mathcal{E}$ stores the optical flow between two keyframes. 
For every incoming frame, we calculate the flow with respect to the last added keyframe in the graph. If the mean flow magnitude is larger than a threshold $\tau\in\mathbb{R}$, it is added to the factor graph and optimized. 

We use the Dense Bundle Adjustment (DBA) layer from~\cite{teed2021droid} to optimize the pose and disparity of the keyframes, as shown in \cref{eq:dba}. This is performed in a sliding window fashion over a local window overlapping the current keyframe and is equivalent to tracking via local BA, as both the camera pose and the geometry are estimated jointly.
\begin{equation}
    \mathop{\arg \min}_{\omega,d} \sum_{(i,j)\in\mathcal{E}} \left\|\Tilde{p}_{ij} - K\omega_j^{-1}(\omega_i(1/d_i)K^{-1}[p_i, 1]^T) \right\|^2_{\Sigma_{ij}}\enspace,
    \label{eq:dba}
\end{equation}
In \cref{eq:dba}, $\Tilde{p}_{ij} \in \mathbb{R}^{(W \times H \times 2) \times 1}$ denotes the flattened predicted pixel coordinates when the pixel grid $p_i \in \mathbb{R}^{(W \times H \times 2) \times 1}$ from keyframe $i$ is projected into keyframe $j$ using optical flow. $d_i \in \mathbb{R}^{(W \times H) \times 1}$ is the disparity map of keyframe $i$, $K$ the camera intrinsics and $\omega_j$ and $\omega_i$ the camera to world extrinsics for keyframes $j$ and $i$ respectively. Finally, $\|\cdot\|_{\Sigma_{ij}}$ denotes the Mahalanobis distance, with $\Sigma_{ij}$ being a diagonal matrix composed of the prediction confidences of the optical flow for each flow term in $\Tilde{p}_{ij}$. Notice that we make two simplifications of the equation: 1. We do not convert the 3D points to homogeneous coordinates and 2. We do not divide by the third coordinate to convert to the pixel space.
 
\boldparagraph{DSPO Layer.}
We find that disparity maps $d$ can still be noisy when only using the DBA layer, which may affect the scale and shift estimation negatively. 
Therefore, we add another optimization objective besides the DBA layer and optimize them alternatively. We call this the Disparity, Scale and Pose Optimization (DSPO) layer which incorporates the predicted monocular depth prior $D^{\text{mono}}$ into the tracking loop. 
We alternatingly optimize the pose and disparity in \cref{eq:dba} and the scale $\theta$, shift $\gamma$ and the high error disparities $d^h$ as
\begin{align}
    \label{eq:dba_wq}
    \mathop{\arg \min}_{d^{h},\theta,\gamma}& \sum_{(i,j)\in\mathcal{E}}\left\|\Tilde{p}_{ij} - K\omega_j^{-1}(\omega_i(1/d_i^h)K^{-1}[p_i, 1]^T) \right\|^2_{\Sigma_{ij}} \\ 
    +& \alpha_1 \sum_{i\in\mathcal{V}}\left\| d_i^h -  \left(\theta_i (1/D^{\text{mono}}_i) + \gamma_i\right)  \right\|^2  
    + \alpha_2 \sum_{i\in\mathcal{V}} \left\| d_i^l -  \left(\theta_i (1/D^{\text{mono}}_i) + \gamma_i\right)  \right\|^2
    \enspace. \nonumber
\end{align}
In \cref{eq:dba_wq}, $d_i^l$ and $d_i^h$ are the valid (low error) and invalid (high error) parts of the disparity map for frame $i$ as defined by the two-view consistency threshold in \cref{eq:two_view_consist}. During optimization, the high error disparity $d^h$ is regularized by the monocular prior, while the scale and shift are refined by the low error disparity $d^l$. We use hyperparameters $\alpha_1 \!<\! \alpha_2$, giving more weight to pixels with accurate disparities ensuring that the scale and shift converge to the right local minima while at the same time optimizing the inaccurate part of the disparity map.
The scale $\theta$ and shift $\gamma$ are initialized as in \cref{eq:wq_least_square}, with the help of point cloud $P_d$.
As also found by \cite{zhang2023hi}, because of scale ambiguity, we do not optimize $d$, $\theta$, $\gamma$ and $\omega$ together, but alternatingly via \cref{eq:dba} and \cref{eq:dba_wq} using the Gauss-Newton algorithm.
Contrary to \cite{zhang2023hi}, we only optimize the high-error disparity pixels, while letting the low-error pixels refine the scale $\theta$ and shift $\gamma$. 
We deploy the DSPO layer for local BA, loop closure and global BA (details below).

\boldparagraph{Loop Closure.} Apart from frame tracking within a local window, we use loop closure and online global BA to avoid scale and pose drift. 
For loop detection, we compute the mean optical flow magnitude between the current active keyframes (within the local window) and all past keyframes. 
For every pair, two conditions are checked: 
First, the flow should fall below a threshold $\tau_{\text{loop}}$ to ensure there is enough co-visibility between the two views. 
Second, the time interval between these two frames need to be larger than a predefined threshold $\tau_{t}$ to avoid redundant edges in the graph. If both conditions are met, a unidirectional edge is added to the graph. During loop closure optimization, only the active keyframes and connected loop nodes are optimized to reduce the computational cost.

\boldparagraph{Global BA.} For online global BA, we build an additional graph containing all keyframes so far. Edges are added based on the temporal and spatial distance between the keyframe pairs as in \cite{zhang2023go}. We found that the open sourced code of \cite{zhang2023go} did not give stable results. To alleviate this, we run online global BA every 20 keyframes. To further improve numerical stability during optimization, we normalize the scale of the disparity maps and poses before each round of global BA. This is done by computing the mean disparity $\bar{d}$ over all keyframes, and updating the disparity as $d_{norm} = d/\bar{d}$ and the pose translation $t$ as $t_{norm} = \bar{d}t$.


\section{Experiments}
\label{sec:exp}


We first describe our experimental setup and then evaluate our method against state-of-the-art dense neural RGB and RGBD SLAM methods on Replica~\cite{straub2019replica} as well as the real world TUM-RGBD~\cite{Sturm2012ASystems} and the ScanNet~\cite{Dai2017ScanNet} datasets. 
For more experiments and details, we refer to the supplementary material.

\boldparagraph{Implementation Details.}
We use $\rho = 0.05$ for the depth interval for both anchoring neural points and rendering. For the search radius parameters, we use $\beta_1 = -0.4$, $\beta_2=0$, and $r_u=0.027$, $r_l = 0.007$ are the upper-bond and lower-bond radii. 
For the proxy depth, we use $\eta=0.01$ to filter points which are inconsistent between view pairs, and use the condition $n_{c}\geq2$ to ensure multi-view consistency. For the mapping loss function, we use $\lambda_\text{geo}=1.0$, $\lambda_\text{pix}=1000.0$ and $\lambda_\text{color}=0.1$. 
We sample $M=2K$ pixels for each mapping phase in ScanNet and TUM-RGBD, and $M=5K$ in Replica. For the number of selected keyframes during the mapping, we use $\kappa=12$ in Replica, $\kappa=10$ in TUM-RGBD and $\kappa=5$ in ScanNet. 
For tracking, we use $\alpha_1=0.01$ and $\alpha_2=0.1$ as weights for the DSPO layer. We use the flow threshold $\tau = 4.0$ on ScanNet, $\tau=3.0$ on TUM-RGBD and $\tau=2.25$ on Replica. The threshold for loop detection is $\tau_\text{loop} = 25.0$. The time interval threshold is $\tau_t = 20$. We conducted the experiments on a cluster with an 1.50GHz AMD EPYC 7742 64-Core CPU and an NVIDIA GeForce RTX 3090 with 24 GiB GPU memory.

\boldparagraph{Evaluation Metrics.}
We evaluate three SLAM tasks: rendering and reconstruction quality as well as tracking accuracy.
For rendering we report PSNR, SSIM~\cite{wang2004image} and LPIPS~\cite{zhang2018unreasonable} on the rendered keyframe images against the sensor images. 
For reconstruction, we first extract the meshes with marching cubes~\cite{lorensen1987marching} as in~\cite{sandstrom2023point} and evaluate the meshes using accuracy $[cm]$, completion $[cm]$ and completion ratio $[\%]$ against the provided ground truth meshes. 
Further, the depth L1 $[cm]$ evaluates the depth on the mesh at random poses against ground truth as in~\cite{zhu2022nice}. 
We use ATE RMSE $[cm]$~\cite{Sturm2012ASystems} to evaluate the estimated trajectory.

\boldparagraph{Datasets.} 
The Replica dataset~\cite{straub2019replica} consists of high-quality synthetic 3D reconstructions of diverse indoor scenes. We leverage
the publicly available dataset~\cite{Sucar2021IMAP:Real-Time}, which
contains trajectories from an RGBD sensor. Additionally,
we showcase our framework on real-world data using the
TUM-RGBD dataset~\cite{Sturm2012ASystems} and the ScanNet dataset~\cite{Dai2017ScanNet}.
The TUM-RGBD poses were captured utilizing an external motion capture system, while ScanNet uses poses from BundleFusion~\cite{dai2017bundlefusion}.

\boldparagraph{Baseline Methods.} We compare our method to numerous published and concurrent works on dense RGB and RGBD SLAM. Concurrent works are denoted with an asterix\textcolor{red}{$^*$}. The main baselines are GO-SLAM~\cite{zhang2023go} and HI-SLAM~\cite{zhang2023hi}.

\begin{table*}[tb]
    \centering    
    \scriptsize
    \setlength{\tabcolsep}{8.0pt}
    \begin{tabular}{lcccccc}
    \toprule
    Metrics  & \makecell[c]{GO-\\SLAM~\cite{zhang2023go}} & \makecell[c]{NICER-\\SLAM~\cite{zhu2023nicer}} & \makecell[c]{MoD-\\SLAM$\textcolor{red}{^*}$~\cite{li2023dense}} &  \makecell[c]{Photo-\\SLAM$\textcolor{red}{^*}$~\cite{huang2023photo}}& \makebox[0.13\linewidth]{\textbf{Ours}}\\
    \midrule
    PSNR$\uparrow$    & 22.13 & 25.41 & \rd27.31 & \fst33.30 & \nd31.04\\
    SSIM $\uparrow$   &  0.73 & 0.83  & \rd0.85  & \nd0.93  & \fst0.97\\
    LPIPS$\downarrow$ &  -   & \nd0.19  & -     & -     & \fst0.12\\  \midrule
    \makecell[l]{ATE RMSE}$\downarrow$& \nd0.39 & 1.88 & \fst0.35 & \rd1.09 & \fst0.35\\  
    \bottomrule
    \end{tabular}
    \caption{
    \textbf{Rendering and Tracking Results on Replica~\cite{straub2019replica} for RGB-Methods}. Our method outperforms all methods on rendering except the concurrent Photo-SLAM$\textcolor{red}{^*}$, where we see similar performance. Our method achieves the best tracking accuracy among all methods. Results are averaged over 8 scenes. Results are from~\cite{tosi2024nerfs} except ours. The best results are highlighted as \colorbox{colorFst}{\bf first}, \colorbox{colorSnd}{second}, and \colorbox{colorTrd}{third}.
    }
    \label{tab:render_replica}
    \vspace{-2em}
\end{table*}

\begin{table*}[htb]
\centering
\scriptsize
\setlength{\tabcolsep}{7.35pt}
\begin{tabularx}{\linewidth}{llccccccc}
\toprule
Method & Metric & \texttt{0000} & \texttt{0059} & \texttt{0106} & \texttt{0169} & \texttt{0181} & \texttt{0207} & Avg.\\
\midrule
\multicolumn{9}{l}{\cellcolor[HTML]{EEEEEE}{\textit{RGB-D Input}}} \\ 

\multirow{3}{*}{\makecell[l]{NICE-\\SLAM~\cite{zhu2022nice}}}
& PSNR$\uparrow$ & 18.71  & \rd16.55  & \rd17.29  &\nd 18.75 & 15.56 & 18.38 & 17.54\\
& SSIM $\uparrow$ & 0.641 & 0.605 & 0.646 & 0.629 & 0.562 & 0.646 & 0.621\\
& LPIPS$\downarrow$ & 0.561 & 0.534 & \rd0.510 & 0.534 & 0.602 & 0.552 & 0.548\\
[0.8pt] \hdashline \noalign{\vskip 1pt}

\multirow{3}{*}{\makecell[l]{VOX-\\Fusion~\cite{yang2022vox}}}
& PSNR$\uparrow$ & \rd19.06 & 16.38 &\nd 18.46 & \rd18.69 & \rd16.75 & \rd19.66 & \rd18.17\\
& SSIM $\uparrow$ & 0.662 & 0.615 &\nd 0.753 & 0.650 & 0.666 & \rd0.696 & \rd0.673\\
& LPIPS$\downarrow$ & 0.515 & 0.528 &\nd 0.439 & \rd0.513 & 0.532 &\nd 0.500 & \rd0.504\\
[0.8pt] \hdashline \noalign{\vskip 1pt}

\multirow{3}{*}{ESLAM~\cite{mahdi2022eslam}} 
& PSNR$\uparrow$ & 15.70 & 14.48 & 15.44 & 14.56 & 14.22 & 17.32 & 15.29\\
& SSIM $\uparrow$ & \rd0.687 & \rd0.632 & 0.628 & \rd0.656 & \rd0.696 & 0.653 & 0.658\\
& LPIPS$\downarrow$ &\nd 0.449 &\nd 0.450 & 0.529 &\nd 0.486 & \rd0.482 & \rd0.534 &\nd 0.488\\
[0.8pt] \hdashline \noalign{\vskip 1pt}

\multirow{3}{*}{\makecell[l]{Point-\\SLAM~\cite{sandstrom2023point}}}
& PSNR$\uparrow$  &\nd 21.30 &\nd 19.48 & 16.80 & 18.53 &\fst 22.27 &\nd 20.56 &\nd 19.82\\
& SSIM $\uparrow$ &\nd 0.806 &\nd 0.765 & \rd0.676 &\nd 0.686 &\fst 0.823 &\nd 0.750 &\nd 0.751\\
& LPIPS$\downarrow$ &\rd0.485 & \rd0.499 & 0.544 & 0.542 &\nd 0.471 & 0.544 & 0.514\\

\midrule
\multicolumn{9}{l}{\cellcolor[HTML]{EEEEEE}{\textit{RGB Input}}} \\ 

\multirow{3}{*}{\makecell[l]{GO-\\SLAM~\cite{zhang2023go}}} 
& PSNR$\uparrow$ &15.74 & 13.15 & 14.58 & 14.49 & 15.72 & 15.37 & 14.84\\
& SSIM $\uparrow$ &0.423 & 0.315 & 0.459 & 0.416 & 0.531 & 0.385 & 0.421\\
& LPIPS$\downarrow$ &0.614 & 0.598 & 0.594 & 0.574 & 0.615 & 0.602 & 0.599\\
[0.8pt] \hdashline \noalign{\vskip 1pt}

\multirow{3}{*}{\textbf{Ours}} 
& PSNR$\uparrow$ &\fst23.42 &\fst 20.66 &\fst 20.41 &\fst 25.23 &\nd 21.28 &\fst 23.68 &\fst 22.45\\
& SSIM $\uparrow$ &\fst0.867 &\fst 0.828 &\fst 0.840 &\fst 0.912 &\nd 0.759 &\fst 0.850 &\fst 0.842\\
& LPIPS$\downarrow$ &\fst0.262 &\fst 0.306 &\fst 0.313 &\fst 0.214 &\fst 0.439 &\fst 0.287 &\fst 0.304\\
\bottomrule
\end{tabularx}

\caption{\textbf{Rendering Performance on ScanNet~\cite{Dai2017ScanNet}.} Our method performs even better than all RGB-D methods on commonly reported rendering metrics. For all RGB-D methods, we take the numbers from~\cite{yugay2023gaussianslam}.}
\label{tab:render_scannet}
\vspace{-2em}
\end{table*}

\begin{figure}[ht]
\vspace{0em}
\centering
{
\setlength{\tabcolsep}{1pt}
\renewcommand{\arraystretch}{1}
\newcommand{\sz}{0.23}
\newcommand{\subsz}{0.2}
\begin{tabular}{ccccc}
\rotatebox{90}{\tiny \makecell{GO-\\\hspace{0.5em}SLAM~\cite{zhang2023go}}}&
\includegraphics[width=\sz\linewidth]{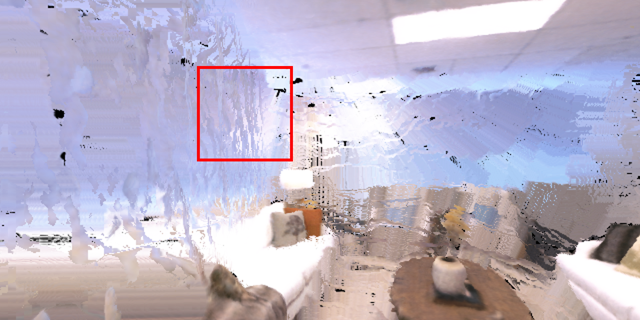} &
\includegraphics[width=\sz\linewidth]{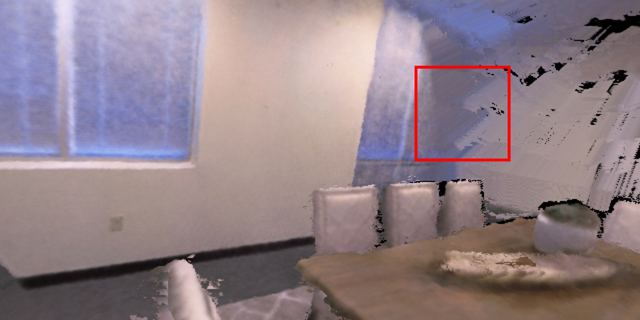} &
\includegraphics[width=\sz\linewidth]{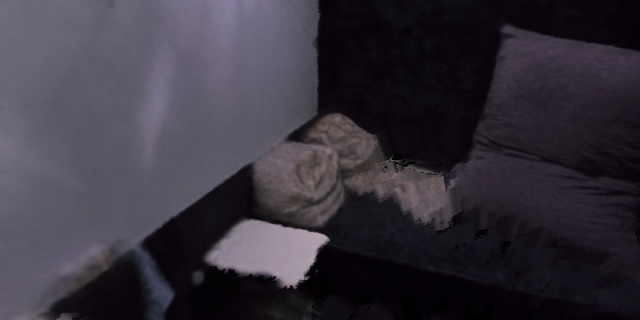} &
\includegraphics[width=\sz\linewidth]{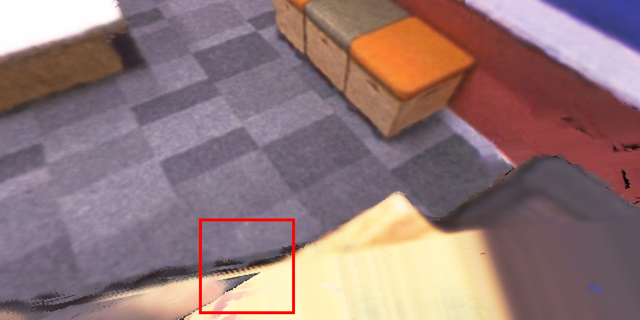} 
\\
\rotatebox{90}{ \tiny \textbf{\makecell{GlORIE-\\SLAM (ours)}}}&
\includegraphics[width=\sz\linewidth]{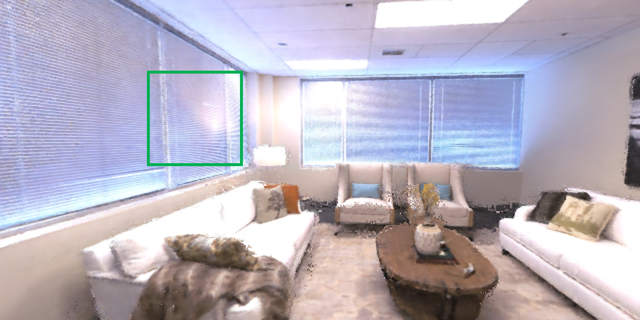} &
\includegraphics[width=\sz\linewidth]{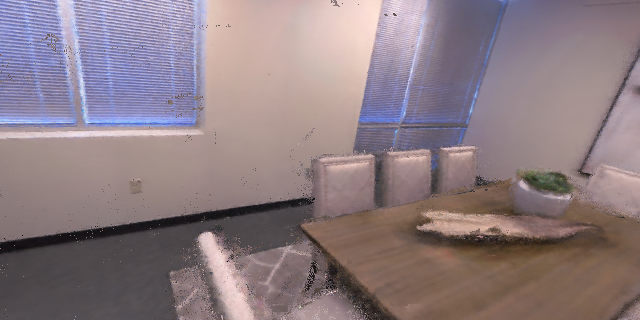} &
\includegraphics[width=\sz\linewidth]{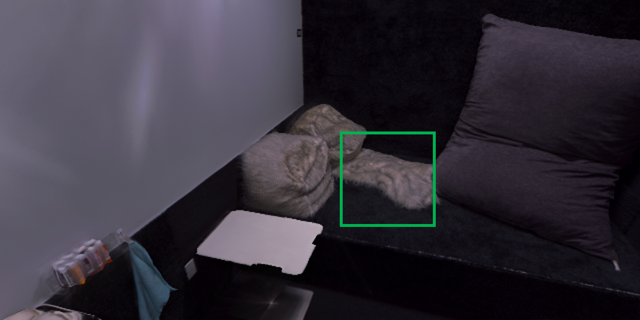} &
\includegraphics[width=\sz\linewidth]{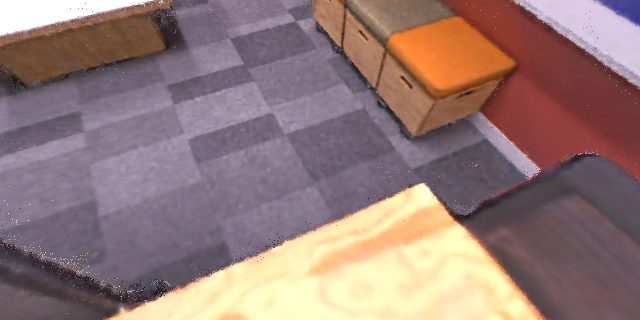} 
\\[0.5em]
\rotatebox{90}{\tiny \makecell{\hspace{1em}Ground\\\hspace{1em}Truth}}&
\includegraphics[width=\sz\linewidth]{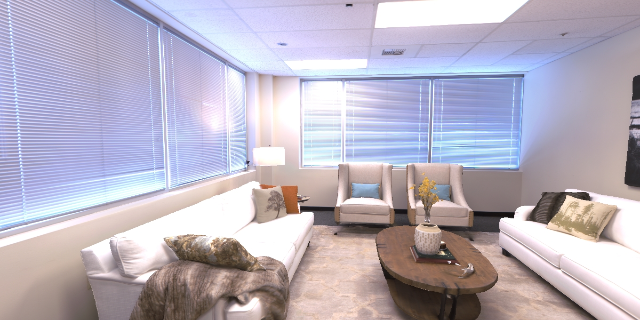} &
\includegraphics[width=\sz\linewidth]{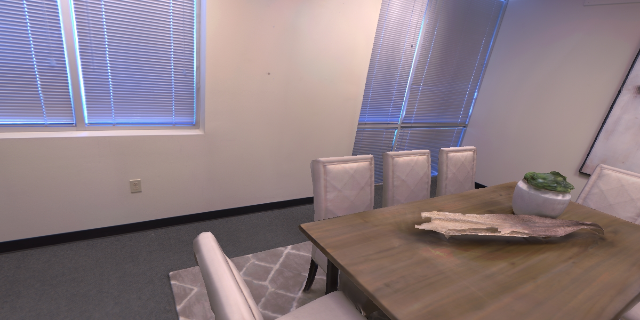} &
\includegraphics[width=\sz\linewidth]{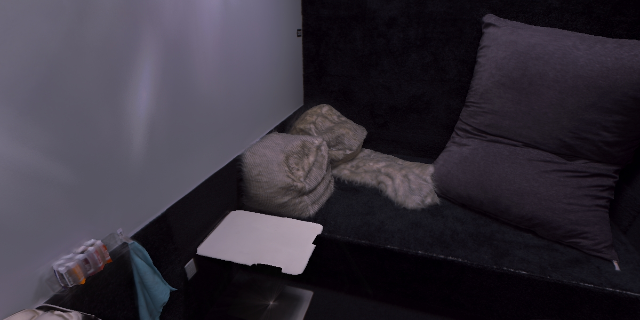} &
\includegraphics[width=\sz\linewidth]{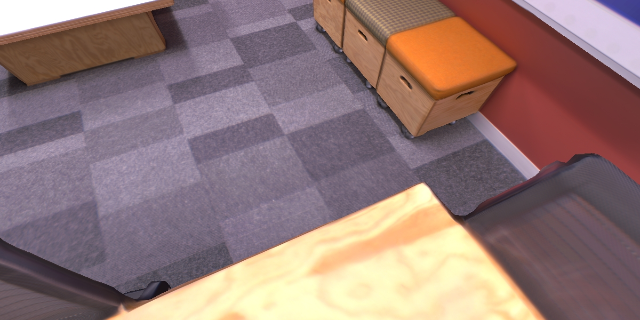} 
\\

\Large
&\texttt{room0}  & \texttt{room2} & \texttt{office1}  & \texttt{office2} \\
\end{tabular}
}
\caption{\textbf{Rendering Results on Replica~\cite{straub2019replica}.} The red boxes show blurry artifacts from GO-SLAM because of insufficient optimization when camera poses and depth are updated. Ours does not suffer from that by deforming the points accordingly. The green boxes show that ours can render high-frequency details well.}
\label{fig:render_replica}
\vspace{0em}
\end{figure}

\begin{figure}[!htb]
\centering
{\scriptsize
\setlength{\tabcolsep}{0pt}
\renewcommand{\arraystretch}{1}
\newcommand{\sz}{0.9}
\begin{tabular}{cc}

\rotatebox{90}{\makecell{GO-\\SLAM~\cite{zhang2023go}}}&\includegraphics[width=\sz\linewidth]{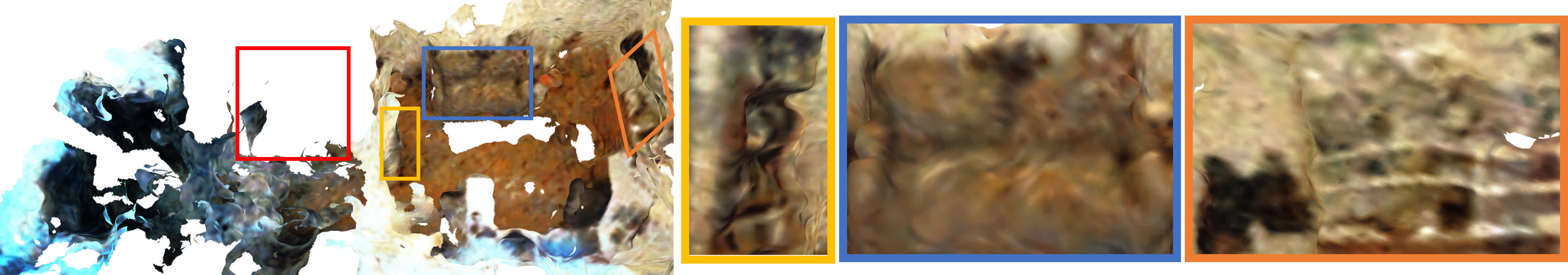} \\ [1.3pt]
\rotatebox{90}{\textbf{\makecell{GlORIE-\\SLAM (ours)}}}&\includegraphics[width=\sz\linewidth]{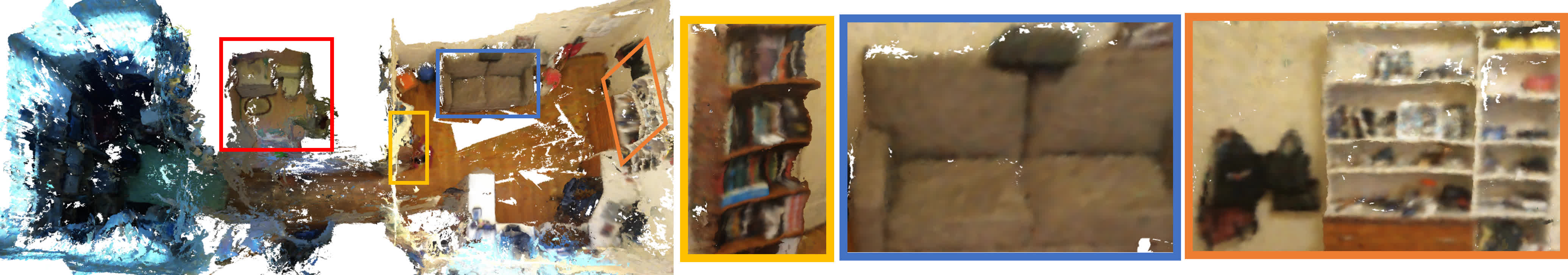} \\ [1.3pt]
\rotatebox{90}{\makecell{\hspace{1.7em}Ground\\\hspace{1.7em}Truth}}&\includegraphics[width=\sz\linewidth]{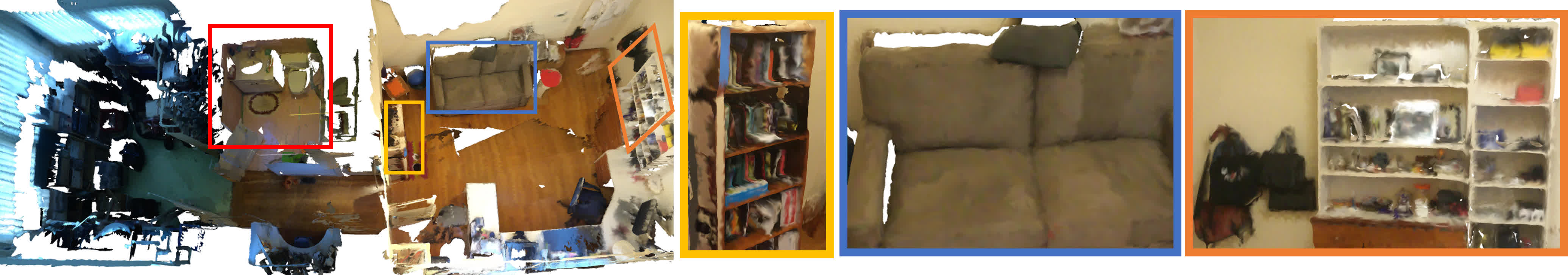} \\ 

& \footnotesize Textured Mesh \qquad \qquad \qquad \qquad Zoom-in Details of the Mesh
\end{tabular}
}
\caption{\textbf{Reconstruction Results on ScanNet \texttt{scene 0054}.} The red box shows the bathroom which GO-SLAM fails to reconstruct while ours succeeds. The yellow, blue and orange boxes show that our method can produce more detailed reconstructions.}
\label{fig:recon_scannet}
\vspace{-0.3em}
\end{figure}

\boldparagraph{Rendering.}
In \cref{tab:render_replica}, we evaluate the rendering performance on Replica~\cite{straub2019replica}. 
We outperform all published works, while beating the concurrent MoD-SLAM$\textcolor{red}{^*}$\cite{zhou2024modslam} and being competitive with Photo-SLAM$\textcolor{red}{^*}$\cite{huang2023photo}, based on Gaussian splatting. 
A qualitative evaluation is shown in \cref{fig:render_replica}. Compared to GO-SLAM~\cite{zhang2023go}, our method does not suffer from blurry artifacts and can even render high-frequency details well, \eg the blinds in the first and second column. 
We further show competitive performance on the real-world dataset ScanNet~\cite{Dai2017ScanNet} in \cref{tab:render_scannet}. 
Our method performs the best, including RGB-D methods. We attribute this to our deformable point cloud scene representation, which allows to be updated without further optimization when the estimated camera trajectory is globally refined. 
None of the RGB-D methods employ loop closure and suffer from pose drift, resulting in worse rendering performance. 
GO-SLAM, which is globally optimized requires feature refinement every time a global update is done, meaning that the optimization never settles, leading to the worst performance on all metrics.

\begin{table*}[t]
    \centering
    \scriptsize
    \setlength{\tabcolsep}{2.2pt}
    \begin{tabular}{lccccccc}
    \toprule
    Metrics  & \makecell[l]{NeRF-\\SLAM~\cite{tosi2024nerfs}} & \makecell[l]{DIM-\\SLAM~\cite{li2023dense}} & \makecell[l]{GO-\\SLAM~\cite{zhang2023go}} &  \makecell[l]{NICER-\\SLAM~\cite{zhu2023nicer}}&  \makecell[l]{HI-\\SLAM~\cite{zhang2023hi}} & \makecell[l]{MoD-\\SLAM$\textcolor{red}{^*}$\!\cite{zhou2024modslam}} & \makebox[0.1\linewidth]{\textbf{Ours}}\\
    \midrule
    \makecell[l]{Depth L1}$\downarrow$& 4.49 & - & {4.39} & - & \rd{3.63} & \fst3.23 & \nd{3.24}\\
    Accuracy $\downarrow$             & - & 4.03 & 3.81 & {3.65} & \rd{3.62} & \fst2.48 & \nd{2.96}\\
    Completion $\downarrow$           & - & \rd{4.20} & 4.79 & \nd{4.16} & 4.59 & - & \fst{3.95}\\ 
    Comp. Rat. $\uparrow$             & - & \rd{79.60} & 78.00 & 79.37& \nd{80.60} & - & \fst{83.72}\\
    \bottomrule
    \end{tabular}
    \caption{
    \textbf{Reconstruction Results on Replica~\cite{straub2019replica} for RGB-Methods.} Our method achieves the best reconstruction performance on all metrics compared to published works. We perform similarly to the concurrent MoD-SLAM$\textcolor{red}{^*}$, which finetunes a mono-depth estimator, while ours does not. Results are averaged over 8 scenes.
    }
    \label{tab:recon_replica}
    \vspace{-2em}
\end{table*}

\boldparagraph{Reconstruction.}
We report reconstruction metrics on Replica in \cref{tab:recon_replica}. 
Our method achieves the best performance on completion and completion ratio, for depth L1 and accuracy ours also achieves competitive results with MoD-SLAM$\textcolor{red}{^*}$~\cite{zhou2024modslam}, which finetunes mono-depth estimator and does extra training while ours does not. 
\cref{fig:recon_scannet} shows qualitative results on a ScanNet multi-room scene.

\begin{table}[htb]
    \centering
    \scriptsize
    \setlength{\tabcolsep}{5.4pt}
    \begin{tabular}{lcccccccccc}
    \toprule
    Method  & \texttt{0000} & \texttt{0059} & \texttt{0106} & \texttt{0169}& \texttt{0181} & \texttt{0207} & \textbf{Avg.-6} & \texttt{0054} & \texttt{0233} & \textbf{Avg.-8}\\
    \midrule
    \multicolumn{11}{l}{\cellcolor[HTML]{EEEEEE}{\textit{RGB-D Input}}} \\ 
    NICE-SLAM~\cite{zhu2022nice}& 12.0 & 14.0 & 7.9 & 10.9 & 13.4 & \nd 6.2 & 10.7 & \rd 20.9 & 9.0 & 11.8 \\
    Co-SLAM~\cite{Wang_2023_CVPR} & 7.1 & 11.1 & 9.4 & \fst 5.9 & 11.8 & 7.1 & 8.7 & - & - & - \\
    ESLAM~\cite{mahdi2022eslam} & 7.3 & \rd 8.5 & \rd 7.5 & \nd 6.5 & \rd 9.0 & \fst 5.7 & \fst 7.4 & 36.3 & \fst 4.3 & \rd 10.6\\
    \midrule
    \multicolumn{11}{l}{\cellcolor[HTML]{EEEEEE}{\textit{RGB Input}}} \\ 
    GO-SLAM~\cite{zhang2023go}  & \nd 5.9 & \nd 8.3 & 8.1 & 8.4 & \nd 8.3 & \rd 6.9 & \rd 7.7 & \nd 13.3 & \rd 5.3 & \nd 8.1\\
    HI-SLAM\cite{zhang2023hi}   & \rd 6.4 & \fst 7.2 & \fst 6.5 & 8.5 & \fst 7.6 & 8.4 & \fst 7.4 & - & -  & -\\
    \textbf{Ours} & \fst 5.5 & 9.1 & \nd 7.0 & \rd 8.2 & \nd 8.3 & 7.5 & \nd 7.6 & \fst 9.4 & \nd 5.1 & \fst 7.5 \\
    \bottomrule
    \end{tabular}
    \caption{
    \textbf{Tracking Accuracy ATE RMSE [cm] $\downarrow$ on ScanNet~\cite{Dai2017ScanNet}.} Our method performs on average competitively with HI-SLAM and better than all other methods. Results for the RGB-D methods are from \cite{liso2024loopyslam}.
    }
    \label{tab:track_scannet}
    \vspace{-2em}
\end{table}

\begin{table}[!htb]
    \centering
    \scriptsize
    \setlength{\tabcolsep}{6.5pt}
    \begin{tabularx}{\linewidth}{lccccccc}
    \toprule
    Method  & \texttt{f1/desk} & \texttt{f2/xyz} & \texttt{f3/off}  & \textbf{Avg.-3} & \texttt{f1/desk2} & \texttt{f1/room}& \textbf{Avg.-5}\\
    \midrule
    \multicolumn{8}{l}{\cellcolor[HTML]{EEEEEE}{\textit{RGB-D Input}}} \\ 
    SplaTAM$\textcolor{red}{^*}$~\cite{keetha2023splatam}& 3.4 & 1.2 & 5.2 & 3.3 & \nd6.5 & 11.1 & \rd5.5\\
    GS-SLAM$\textcolor{red}{^*}$~\cite{yan2023gs} & \fst1.5 & 1.6 & 1.7 &1.6 & - & - & -\\
    GO-SLAM~\cite{zhang2023go} & \fst1.5 & \nd0.6 & \nd1.3 &\fst1.1 & - & \nd 4.7 & -\\
    \midrule
    \multicolumn{8}{l}{\cellcolor[HTML]{EEEEEE}{\textit{RGB Input}}} \\ 
    MonoGS$\textcolor{red}{^*}$~\cite{matsuki2023gaussian} & 4.2 & 4.8 & 4.4 & 4.4 & - & - & - \\
    Photo-SLAM$\textcolor{red}{^*}$~\cite{huang2023photo} & \fst1.5 & 1.0 & \nd1.3 & \rd1.3 & - & - & -  \\
    DIM-SLAM~\cite{li2023dense}  & \rd2.0 & \nd0.6 & 2.3 & 1.6 & - & - & - \\
    GO-SLAM~\cite{zhang2023go} & \nd1.6 & \nd0.6 &1.5 & \nd1.2 & \fst2.8 & \rd5.2 & \nd2.3 \\
    MoD-SLAM$\textcolor{red}{^*}$~\cite{zhou2024modslam} & \fst1.5 & \rd0.7 & \fst1.1 & \fst1.1 & - & - & - \\
    \textbf{Ours} & \nd1.6 & \fst0.2 & \rd1.4 & \fst1.1 & \fst2.8 & \fst4.2 & \fst2.1 \\

    \bottomrule
    \end{tabularx}
    \caption{
    \textbf{Tracking Accuracy ATE RMSE [cm] $\downarrow$ on TUM-RGBD~\cite{Sturm2012ASystems}}. Our method performs on average the best among all the methods. Note that all methods with a $\textcolor{red}{^*}$ are concurrent works.
    }
    \label{tab:track_tum_rgbd}
    \vspace{-2em}
\end{table}

\boldparagraph{Tracking.}
In \cref{tab:render_replica}, \cref{tab:track_scannet} and \cref{tab:track_tum_rgbd}, we report the tracking accuracy of the estimated trajectory on Replica~\cite{straub2019replica}, ScanNet~\cite{Dai2017ScanNet} and TUM-RGBD~\cite{Sturm2012ASystems}. On all three datasets, our method shows competitive results in every single scene and gives the best average value among the RGB and RGB-D methods.

\begin{figure}[t]
\centering
{
\footnotesize
\setlength{\tabcolsep}{0.5pt}
\renewcommand{\arraystretch}{1}
\newcommand{\sz}{0.197}
\begin{tabular}{ccccc}
\includegraphics[width=\sz\linewidth]{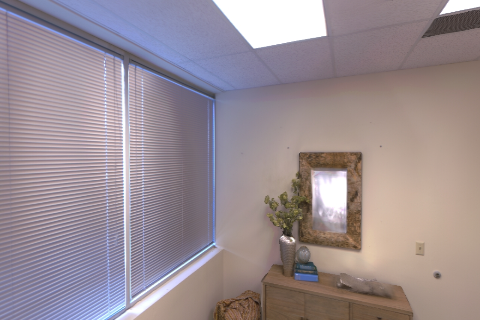} &
\includegraphics[width=\sz\linewidth]{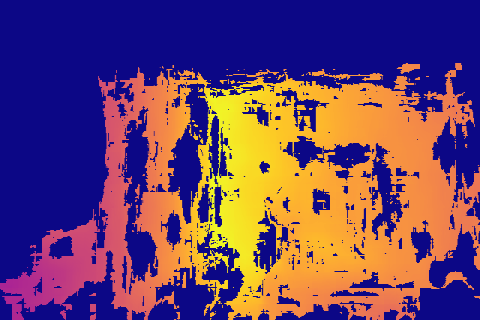} &
\includegraphics[width=\sz\linewidth]{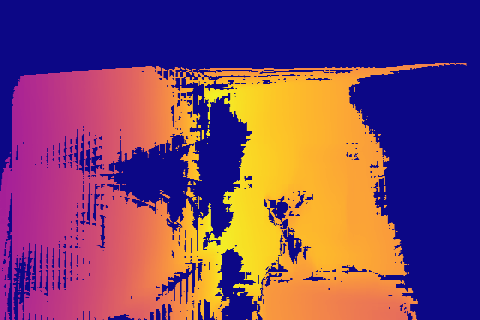} &
\includegraphics[width=\sz\linewidth]{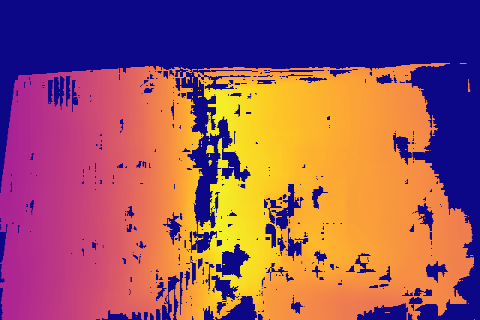} &
\includegraphics[width=\sz\linewidth]{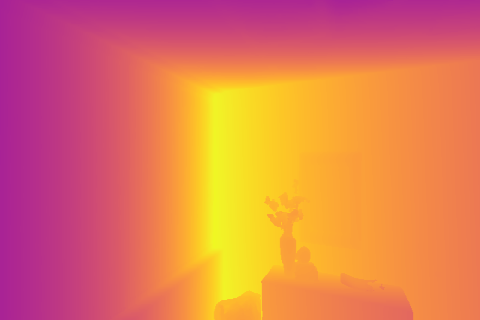} \\[-2pt]
 \texttt{RGB} & \texttt{DBA}  & \texttt{DBA+mono prior} & \texttt{\textbf{DSPO}}  & \texttt{Ground Truth} \\[-3pt]
\end{tabular}
}
\caption{\textbf{Comparison of Estimated Depth.} We show the depth output $\Tilde{D}$ from the tracker. The pixels which are invalid (high error) are colored dark blue. \texttt{DBA} is the method that Droid-SLAM~\cite{teed2021droid} uses. The \texttt{DBA+mono prior} strategy is used in HI-SLAM~\cite{zhang2023hi}, \ie the mono prior supervises all pixels directly. It is clear that our formulation (\texttt{DSPO}) provides the most consistent keyframe depth.}
\vspace{-0em}
\label{fig:BaD}
\end{figure}

\begin{table}[htb]
    \centering
    \scriptsize
    \setlength{\tabcolsep}{5.5pt}
    \begin{tabularx}{\linewidth}{cccccccc}
    \toprule
    \makecell[c]{Deformable\\Pointcloud}&\makecell[c]{Mono Prior\\ in Proxy Depth }&\makecell[c]{DSPO\\Layer} &\makecell[c]{PSNR\\$[dB]\uparrow$} & \makecell[c]{Depth L1\\$[cm]\downarrow$} &\makecell[c]{Acc.\\$[cm]\downarrow$} &\makecell[c]{Comp.\\$[cm]\downarrow$}& \makecell[c]{Comp. Ratio\\$[\%]\uparrow$}\\
    \midrule
    \redx & \redx & \redx & {\cellcolor[HTML]{EEEEEE}{25.52}} & {\cellcolor[HTML]{EEEEEE}{9.21}} & {\cellcolor[HTML]{EEEEEE}{5.91}} &{\cellcolor[HTML]{EEEEEE}{7.70}} &{\cellcolor[HTML]{EEEEEE}{73.02}}\\
    \redx & \greencheck & \redx &{\cellcolor[HTML]{EEEEEE}{25.58}} & {\cellcolor[HTML]{EEEEEE}{9.07}} & {\cellcolor[HTML]{EEEEEE}{5.76}} &{\cellcolor[HTML]{EEEEEE}{6.61}} &{\cellcolor[HTML]{EEEEEE}{76.78}}\\
    \greencheck & \redx &\redx &{\cellcolor[HTML]{EEEEEE}{30.89}} & {\cellcolor[HTML]{EEEEEE}{3.55}} & {\cellcolor[HTML]{EEEEEE}{3.61}} &{\cellcolor[HTML]{EEEEEE}{5.19}} &{\cellcolor[HTML]{EEEEEE}{78.84}}\\
    
    \greencheck & \greencheck & \redx &{\cellcolor[HTML]{EEEEEE}{30.66}} & {\cellcolor[HTML]{EEEEEE}{3.52}} & {\cellcolor[HTML]{EEEEEE}{3.72}} & {\cellcolor[HTML]{EEEEEE}{\textbf{3.92}}}&{\cellcolor[HTML]{EEEEEE}{83.40}}\\
    \greencheck & \greencheck &\greencheck &{\cellcolor[HTML]{EEEEEE}{\textbf{31.04}}} & {\cellcolor[HTML]{EEEEEE}{\textbf{3.24}}} & {\cellcolor[HTML]{EEEEEE}{\textbf{2.96}}} &{\cellcolor[HTML]{EEEEEE}{3.95}} &{\cellcolor[HTML]{EEEEEE}{\textbf{83.72}}}\\
           
    \bottomrule
    \end{tabularx}
    \caption{
    \textbf{Ablation Study on Replica~\cite{straub2019replica}.} The first 4 rows use DBA~\cite{teed2021droid} instead of DSPO. Deforming the neural point cloud is important for reconstruction and rendering, the monocular prior helps with completeness and the DSPO layer improves both rendering and reconstruction metrics. All results are averaged over 8 scenes.
    }
    \label{tab:ablation}
    \vspace{-2em}
\end{table}

\boldparagraph{Ablation Study.}
In \cref{tab:ablation}, we conduct a set of ablation studies related to our method, by enabling and disabling certain parts. We find that deforming the neural point cloud is important for rendering and geometry prediction. The monocular depth prior is important for the geometric completeness and the DSPO layer helps improve both the rendering and reconstruction accuracy. In \cref{fig:BaD}, we compare the valid estimated depth $\Tilde{D}$, showing that our proposed DSPO layer contributes significantly to more consistent depth estimation.

\begin{table}[!htb]
    \centering
    \scriptsize
    \setlength{\tabcolsep}{11pt}
    \begin{tabular}{lcccc}
    \toprule
       & Point-SLAM~\cite{sandstrom2023point} & GO-SLAM~\cite{zhang2023go} & SplaTAM$\textcolor{red}{^*}$\cite{keetha2023splatam} &Ours\\
    \midrule
    GPU Mem [GiB] & {\cellcolor[HTML]{EEEEEE}{7.11}} & {\cellcolor[HTML]{EEEEEE}{18.50}} & {\cellcolor[HTML]{EEEEEE}{18.54}} & {\cellcolor[HTML]{EEEEEE}{15.22}}\\ 
    \midrule
    Avg. FPS & {\cellcolor[HTML]{EEEEEE}{0.23}}  & {\cellcolor[HTML]{EEEEEE}{8.36}}& {\cellcolor[HTML]{EEEEEE}{0.14}} & {\cellcolor[HTML]{EEEEEE}{0.23}}\\
    \bottomrule
    \end{tabular}
    \caption{
    \textbf{Memory and Running Time Evaluation on Replica~\cite{straub2019replica} \texttt{room0}.} Our peak memory usage and runtime is comparable to existing works. We take the numbers from \cite{tosi2024nerfs} except for ours. All methods except for ours are evaluated in RGB-D mode. All methods are evaluated using an NVIDIA RTX 3090 GPU. 
    }
    \label{tab:mem_and_time}
    \vspace{-2em}
\end{table}

\boldparagraph{Memory and Running Time.}
In \cref{tab:mem_and_time}, we evaluate the peak GPU memory usage and runtime of our method. We achieve a comparable memory usage with GO-SLAM~\cite{zhang2023go} which is the most representative method using frame-to-frame tracking. SplaTAM~\cite{keetha2023splatam}, which is a RGB-D 3D GS method has a similar memory footprint. We use more memory compared to the RGB-D method Point-SLAM~\cite{sandstrom2023point}. Although we both use the neural point cloud representation, we need to add more neural points to the scene to resolve the noisy depth, which Point-SLAM can avoid by having access to ground truth depth.
Regarding runtime, we are comparable with Point-SLAM and SplaTAM. GO-SLAM has the fastest runtime, but as shown in \cref{tab:render_replica} and \cref{tab:recon_replica}, it sacrifices rendering and reconstruction quality for speed.  

\boldparagraph{Limitations.} We currently do not prune neural points that may be redundant or wrongly positioned as a result of depth noise. Most points are, however, corrected by the point cloud deformations, but point pruning would nonetheless be useful. Another limitation is that our construction of the final proxy depth $D_c$ is rather simple and does not fuse the monocular and keyframe depth maps in an informed manner, by utilizing \eg normal consistency. Finally, as future work, it is interesting to study how and when the frame-to-frame tracking paradigm can be boosted by frame-to-model techniques.

\section{Conclusion}
\label{sec:conclusion}
We proposed \ours, a dense RGB-only SLAM system which utilizes a deformable neural point cloud for mapping and a globally optimized frame-to-frame tracking technique driven by optical flow.
Importantly, the inclusion of a monocular depth prior into the tracking loop, to refine the scale and to regularize the erroneous keyframe depth predictions, leads to better rendering and mapping. 
By using the monocular prior for depth completion, mapping is further improved. 
Our experiments demonstrate that \ours outperforms existing solutions regarding reconstruction and rendering accuracy while being on par or better with respect to tracking as well as runtime and memory usage.

\vspace{3pt}
\boldparagraph{Acknowledgments.}
We thank Wei Zhang for fruitful discussions.

\clearpage
\title{GlORIE-SLAM: Globally Optimized RGB-only Implicit Encoding Point Cloud SLAM \\ \vspace{0.5em}\large Supplementary Material}
\titlerunning{GlORIE-SLAM Supplementary}
\authorinfo
\maketitle
\appendix
%
\setcounter{table}{7}
\setcounter{equation}{15}
\setcounter{figure}{5}

\label{sec:supp}


\section{Method}

\subsection{Mathematical Derivation of the DSPO Layer}
In the main paper, we introduce the DSPO layer, which optimizes the disparity, scale and pose together. Here, we provide the detailed derivation of how to optimize the proposed DSPO objective using the Gauss-Newton algorithm with the Schur Complement.

As described in the main paper, we optimize the following two objective functions alternatingly. The first one (same as the DBA layer in~\cite{teed2021droid}) optimizes the poses and disparity maps of the involved keyframes,
\begin{equation}
    \mathop{\arg \min}_{\omega,d} \sum_{(i,j)\in\mathcal{E}} \left\|\Tilde{p}_{ij} - K\omega_j^{-1}(\omega_i(1/d_i)K^{-1}[p_i, 1]^T) \right\|^2_{\Sigma_{ij}}\enspace.
    \label{eq:supp_dba}
\end{equation}
The second objective optimizes the scale and shift factors of the mono prior and also the disparity maps of the keyframes,
\begin{align}
    \label{eq:supp_dba_wq}
    \mathop{\arg \min}_{d^{h},\theta,\gamma}& \sum_{(i,j)\in\mathcal{E}}\left\|\Tilde{p}_{ij} - K\omega_j^{-1}(\omega_i(1/d_i^h)K^{-1}[p_i, 1]^T) \right\|^2_{\Sigma_{ij}} \\ 
    +& \alpha_1 \sum_{i\in\mathcal{V}}\left\| d_i^h -  \left(\theta_i (1/D^{\text{mono}}_i) + \gamma_i\right)  \right\|^2  
    + \alpha_2 \sum_{i\in\mathcal{V}} \left\| d_i^l -  \left(\theta_i (1/D^{\text{mono}}_i) + \gamma_i\right)  \right\|^2
    \enspace. \nonumber
\end{align}
For better readability, we do not show the conversion of 3D points to homogeneous coordinates in all equations.

\boldparagraph{DBA Optimization.}
First, we introduce how to solve the DBA optimization. 
To keep the notation consistent with~\cite{teed2022optimization}, we do a mapping of the symbols we use in \cref{eq:supp_dba} such that
\begin{equation}
\{(p_i,\Tilde{p}_{i,j}) \,|\, (i,j)\in \mathcal{E}\}  \rightarrow \mathcal{M}=\{(x^k_i,x^k_j)\}_{k=1}^M \enspace,    
\end{equation}
where $(x^k_i,x^k_j)$ is a corresponding pixel pair (a feature match found by the optical flow estimation), $k$ is the index in the total set of matches $\mathcal{M}$. For each $k$ there are specific indices $i$ and $j$ indicating the corresponding keyframes of $x^k_i$ and $x^k_j$, \ie $i$ and $j$ are functions of $k$, but we write them as a subscript directly for simplicity. Furthermore, we map the camera poses
\begin{equation}
\{\omega_i\}_{i=1}^N \rightarrow \textbf{T}=\{T_i\in SE(3) \;|\; 1\leq i\leq N\} \enspace,
\end{equation}
where $\textbf{T}$ are the camera to world poses of the corresponding keyframes.
Then we can rewrite the objective function in \cref{eq:supp_dba} as
\begin{equation}
    f(\textbf{T},\textbf{d}) = \frac{1}{2} \sum_{(x^k_i,x^k_j)\in \mathcal{M}}\left\|r_k(T_i,T_j,x^k_i,x^k_j,d^k_i) \right\|^2_{w_k}\enspace,
    \label{eq:supp_reproj}
\end{equation}
where $\textbf{d}=\{d_i^k\}_{k=1}^M$ are the disparities (\ie inverse depth), $d_i^k$ is the disparity of pixel $x_i^k$, where $i$ is still a function of $k$, but we write it as a subscript for simplicity. We denote $\|\cdot\|_{w_k}$ as the Mahalanobis distance with weighting matrix $w_k$. $w_k = \text{diag}(w_k^1, w_k^2)$ is a $2 \times 2$ diagonal matrix of the per-pixel and per-direction uncertainties of the optical flow prediction. We add $1/2$ to the objective function for mathematical convenience as it does not change the optimal solution. $r_k(\cdot) \in \mathbb{R}^2$ is the reprojection error function and it is defined as 
\begin{equation}
r_k(T_i,T_j,x^k_i,x^k_j,d^k_i) = x_{j}^k - KT_j^{-1}T_i(1/d_i^k)K^{-1}[x_i^k, 1]^T \enspace.
\label{eq:per_pixel_objective}
\end{equation}
The formulation in \cref{eq:per_pixel_objective} is the per-pixel equivalent of the part inside the norm $\|\cdot\|$ in \cref{eq:supp_dba}. 

To minimize \cref{eq:supp_reproj}, we resort to the Gauss-Newton algorithm. We can define the model parameters by the column vector $\beta = [T_j, T_i, d_i^k]^T \in \mathbb{R}^{13}$. We parameterize rotations and translations with the lie algebra se(3), so each camera pose is parameterized by 6 values and the $d_i^k \in \mathbb{R}$. Despite being an RGB-only system, optimization is done on se(3) because we fix the poses of the two first keyframes after initialization. This prevents gauge freedom \ie drift in the global scale of the scene during optimization and therefore, se(3) optimization is sufficient. 
We can define the model as   
\begin{equation}
\renewcommand\arraystretch{1.5}
g_k(\beta) = \begin{bmatrix}
        g^1_k(\beta) \\ g^2_k(\beta) 
    \end{bmatrix} = KT_j^{-1}T_i(1/d_i^k)K^{-1}[x_i^k, 1]^T \enspace.
\end{equation}
since $g_k \in \mathbb{R}^2$ is not linear with respect to the parameters $\beta$ we approximate $g_k$ around a current estimate $\beta_0$ (at time 0) of the parameters with a first-order Taylor expansion. For each dimension of $g_k$, we get (here for the first dimension)
\begin{equation}
g^1_k(\beta) \approx g^1_k(\beta_0) + J_k(\beta - \beta_0)
\label{eq:taylor_approx}
\end{equation}
where $J_k \in \mathbb{R}^{13}$ is the gradient row vector of $g_k$ with respect to $\beta$ at $\beta_0$. Plugging back \cref{eq:taylor_approx} into \cref{eq:per_pixel_objective}, we get (for the first dimension)
\begin{equation}
r^1_k(\beta) \approx \{x_j^k\}^1 - g^1_k(\beta_0) - J_k(\beta - \beta_0) = r^1_k(\beta_0) - J_k(\beta - \beta_0) \enspace,
\end{equation}
where $\{x_j^k\}^1$ denotes the first dimension of $x_j^k$. The goal of the optimization is to minimize the squared residuals, \ie for pixel $k$ and dimension 1, the term $(r_k^1)^2$. We differentiate $(r_k^1)^2$ with respect to $\beta$ and set the derivative to zero. This yields the expression
\begin{equation}
J_k^TJ_k(\beta - \beta_0) + J_k^Tr_k^1(\beta_0) = 0.
\end{equation}
Now that we have derived the equation to solve (for $\beta$) for a single pixel and dimension, we can combine all observations from all pixels (further details can be found in section 2.4 in~\cite{teed2022optimization}). We achieve this by stacking all residuals as a column vector $\textbf{r}=(r_1^1, r_1^2, \dotsi,r_M^1, r_M^2)^T \in \mathbb{R}^{2M}$. 
This yields the equivalent, vector-valued equation

\begin{align}
    \textbf{J}^T\textbf{W}\textbf{J} (\boldsymbol{\beta} - \boldsymbol{\beta_0}) + \textbf{J}^T\textbf{W}\textbf{r} &= 
    \textbf{J}^T\textbf{W}\textbf{J} \Delta \boldsymbol{\beta}
    + \textbf{J}^T\textbf{W}\textbf{r} \notag\\
    &= 
    \textbf{J}^T\textbf{W}\textbf{J} 
    \begin{bmatrix}
        \Delta \textbf{T} \\ \Delta \textbf{d} 
    \end{bmatrix}
    + \textbf{J}^T\textbf{W}\textbf{r} = 0 \enspace.
    \label{eq:gauss-newton}
\end{align}
Here, we combine the uncertainties into a diagonal matrix as 
\begin{equation}
    \textbf{W}=\text{diag}(w_1^1, w_1^2, \dotsi, w_M^1, w_M^2)\enspace.
\end{equation}

The full Jacobian $\textbf{J}$ has the shape $[2M \times (6N + M)]$, $\Delta \textbf{T}$ is $[6N \times 1]$ and $\Delta \textbf{d}$ is $[M \times 1]$.
To solve \cref{eq:gauss-newton}, we rewrite it as
\begin{equation}
    \textbf{J}^T\textbf{W}\textbf{J} 
    \begin{bmatrix}
        \Delta \textbf{T} \\ \Delta \textbf{d} 
    \end{bmatrix}
    = - \textbf{J}^T\textbf{W}\textbf{r}\enspace.
\end{equation}
We now arrange the full Jacobian matrix $\textbf{J}$ into two blocks as 
%
\begin{equation}
    \textbf{J} = 
    \begin{bmatrix}
        \textbf{J}_\textbf{T} & \textbf{J}_\textbf{d}
    \end{bmatrix}\enspace,
    \label{eq:jacobian}
\end{equation}
where $\textbf{J}_\textbf{T} \in \mathbb{R}^{2M \times 6N}$ is the Jacobian block of $\textbf{r}$ with respect to poses $\textbf{T}$, and $\textbf{J}_\textbf{d} \in \mathbb{R}^{2M \times M}$ is the Jacobian block of $\textbf{r}$ with respect to disparities $\textbf{d}$.
For the detailed derivation of $\textbf{J}_\textbf{T}$ and $\textbf{J}_\textbf{d}$, we refer to section 2.3.1 in~\cite{teed2022optimization}. This yields

\begin{equation}
\renewcommand\arraystretch{1.5}
\setlength{\arraycolsep}{3pt}
    \begin{bmatrix}
        \textbf{J}_\textbf{T}^T \textbf{W} \textbf{J}_\textbf{T} & \textbf{J}_\textbf{T}^T \textbf{W}\textbf{J}_\textbf{d} \\
        \textbf{J}_\textbf{d}^T \textbf{W}\textbf{J}_\textbf{T} & \textbf{J}_\textbf{d}^T \textbf{W}\textbf{J}_\textbf{d} 
    \end{bmatrix}
    \begin{bmatrix}
        \Delta \textbf{T} \\ \Delta \textbf{d} 
    \end{bmatrix}
    = - \textbf{J}^T\textbf{W}\textbf{r}\enspace.
    \label{eq:solve_gn}
\end{equation}
We rewrite \cref{eq:solve_gn} to the following form,
\begin{equation}
\renewcommand\arraystretch{1.5}
\setlength{\arraycolsep}{3pt}
    \begin{bmatrix}
        \textbf{B} & \textbf{E} \\
        \textbf{E}^T & \textbf{C}
    \end{bmatrix}
    \begin{bmatrix}
        \Delta \textbf{T} \\ \Delta \textbf{d} 
    \end{bmatrix}
    =
    \begin{bmatrix}
        \textbf{v} \\ \textbf{w} 
    \end{bmatrix}\enspace,
    \label{eq:schur}
\end{equation}
and use the Schur Complement to solve for $\Delta \textbf{T}$ and $\Delta \textbf{d}$,
\begin{align}
    \Delta\textbf{T}&=\left[\textbf{B}-\textbf{E}\textbf{C}^{-1}\textbf{E}^T\right]^{-1}(\textbf{v}-\textbf{E}\textbf{C}^{-1}\textbf{w})\notag\\
    \Delta\textbf{d}&=\textbf{C}^{-1}(\textbf{w}-\textbf{E}^T\Delta\textbf{T}) \enspace.
\end{align}
Note that $\textbf{C}$ here is diagonal since each $d^k_i$ is only involved in $r_k$. Though $\textbf{J}_{\textbf{d}}$ is not perfectly diagonal (it has two non-zero values along each column), $\textbf{J}_{\textbf{d}}^T \textbf{W}\textbf{J}_{\textbf{d}} = \textbf{C} \in \mathbb{R}^{M \times M}$ is diagonal. Therefore, the inverse of $\textbf{C}$ is easy to compute. Also, the size of $\textbf{B} \in \mathbb{R}^{6N \times 6N}$ is relatively small and therefore it is tractable to compute $\left[\textbf{B}-\textbf{E}\textbf{C}^{-1}\textbf{E}^T\right]^{-1}$. This is achieved by Cholesky decomposition.
Finally we can use 
$\Delta\textbf{T}$ and $\Delta\textbf{d}$ to update the poses and disparities accordingly.

\boldparagraph{Scale, Shift and Disparity Optimization.}
We approach the problem of joint scale, shift and disparity optimization in a similar manner as above. First, we rewrite the objective function in \cref{eq:supp_dba_wq} as,
\begin{align}
    f(\textbf{s},\textbf{d}_h) =  &\dfrac{1}{2} \sum_{\substack{(x^k_i,x^k_j)\in \mathcal{M}_h \\ d_i^k \in \textbf{d}_h }}\left\| r_k(T_i,T_j,x^k_i,x^k_j,d^k_i) \right\|^2_{w_k} + \alpha_1 \left\| t_k(d^k_i,s_i) \right\|^2  \notag\\
    + &\dfrac{1}{2} \sum_{\substack{d_i^k\in\textbf{d}_l }}\alpha_2 \left\|t_k(d^k_i,s_i) \right\|^2
    \label{eq:supp_wq_loss}
\end{align}
where $s_i=(\theta_i,\gamma_i)$ is the scale and shift of frame $i$. We denote the set 
\begin{equation}
    \mathcal{M}_h =\{(x^k_i,x^k_j)\}_{k=1}^H
\end{equation}
as the set of pixels where $d_i^k$ is deemed to have a high error as defined by the multi-view filter in Eq. (\textcolor{red}{6}). The shape of $\textbf{d}_h$ is $[1\times H]$ $(H \leq M)$ and $\textbf{d}_l$ denotes the set of disparities with a low error. Adding the cardinalities of $\textbf{d}_h$ and $\textbf{d}_l$ yields $M$ \ie $| \textbf{d}_h | + | \textbf{d}_l | = H + | \textbf{d}_l | = M$. 
We denote the set of scales and shifts for all frames involves as $\textbf{s}=(s_1, \ldots, s_N)$. 
$t_k(\cdot)\in \mathbb{R}$ is the residual term for the regularization by the monocular depth prior,
\begin{equation}
    t_k(d_i^k,s_i) = d_i^k - (\theta_i\cdot d_\text{mono,i}^k + \gamma_i)\enspace.
\end{equation}
Now, we collect all the residuals as a column vector 
\begin{equation}
    \hat{\textbf{r}} = (\ldots, r_k^1, r_k^2, \ldots, r_H^1, r_H^2, t_1, \ldots, t_M)^T \enspace,
\end{equation}
\ie a collection of all $r_k$ where $k$ needs to ensure $d_i^k$ is invalid (high error) and also all $t$.
To define the corresponding linear system as in \cref{eq:solve_gn}, we begin by defining the full Jacobian matrix $\widehat{\textbf{J}}$ as a collection of two blocks, similar to \cref{eq:jacobian} \ie
$\widehat{\textbf{J}} =
\begin{bmatrix}
    \widehat{\textbf{J}}_\textbf{s} & \widehat{\textbf{J}}_\textbf{d}
\end{bmatrix}
$,
where $\widehat{\textbf{J}}_\textbf{s}\in\mathbb{R}^{(2H+M)\times2N}$ is the Jacobian block of $\widehat{\textbf{r}}$ with respect to the scales and shifts $\textbf{s}$, and $\hat{\textbf{J}}_\textbf{d}\in\mathbb{R}^{(2H+M)\times H}$ is the Jacobian block of $\hat{\textbf{r}}$ with respect to the invalid disparities $\textbf{d}_h$. Here both $\hat{\textbf{J}}_\textbf{s}$ and $\hat{\textbf{J}}_\textbf{d}$ consist of two parts. The first part comes from $r_k(\cdot)$ and second part from $t_k(\cdot)$ \ie,
\begin{equation}
\renewcommand\arraystretch{1.6}
\setlength{\arraycolsep}{3pt}
    \hat{\textbf{J}}_\textbf{s} = 
    \begin{bmatrix}
    \widehat{\textbf{J}}^r_\textbf{s} \\ \widehat{\textbf{J}}^t_\textbf{s}
    \end{bmatrix}
    \qquad
    \hat{\textbf{J}}_\textbf{d} = 
    \begin{bmatrix}
    \widehat{\textbf{J}}^r_\textbf{d} \\ \widehat{\textbf{J}}^t_\textbf{d}
    \end{bmatrix}
    \qquad
    \hat{\textbf{J}} = 
    \begin{bmatrix}
    \widehat{\textbf{J}}^r_\textbf{s} & \widehat{\textbf{J}}^r_\textbf{d}\\
    \widehat{\textbf{J}}^t_\textbf{s} & \widehat{\textbf{J}}^t_\textbf{d}\\
    \end{bmatrix}\enspace.
\end{equation}
$\widehat{\textbf{J}}^r_\textbf{d}$ is the same as ${\textbf{J}}_\textbf{d}$ in \cref{eq:jacobian} except that now it only contains the invalid disparities $\textbf{d}_h$ instead of $\textbf{d}$, \ie the shape of $\widehat{\textbf{J}}^r_\textbf{d}$ is $[2H\times H]$ instead of $[2M\times M]$. Note that we need to use a factor 2 here because each residual $r_k$ has 2 dimensions. $\widehat{\textbf{J}}^r_\textbf{s} \in \mathbb{R}^{2H\times2N}$ is $0$ since none of the $s_i$ are involved in any of the $r_k(\cdot)$ residuals. The derivatives of $t_k(\cdot)$ with respect to $\theta_i$, $\gamma_i$ and $d_i^k$ are,
\begin{equation}
    \dfrac{\partial t_k}{\partial \theta_i} = -d^k_{\text{mono},i} \quad \dfrac{\partial t_k}{\partial \gamma_i} = -1 \quad \dfrac{\partial t_k}{\partial d_i^k} = 1 \enspace.
\end{equation}
Thus the form of $\widehat{\textbf{J}}^t_\textbf{s}\in\mathbb{R}^{M\times2N}$ is as follows,
\begin{equation}
\renewcommand\arraystretch{1.6}
\setlength{\arraycolsep}{3pt}
    \widehat{\textbf{J}}^t_\textbf{s} = 
    \begin{bmatrix}
        J_{1,1} & \dots & J_{1,N} \\
        \vdots & \ddots & \vdots \\
        J_{M,1} &\dots & J_{M,N}
    \end{bmatrix}\quad
    J_{k,i} = \begin{bmatrix}
        -d^k_{\text{mono},i} & -1
    \end{bmatrix}\enspace,
\end{equation}
Finally, we construct $\widehat{\textbf{J}}^t_\textbf{d}$ as follows. First define the diagonal square matrix $\textbf{D} = \text{diag}(1,\ldots,1)\in \mathbb{R}^{M\times M}$. Then we define $\widehat{\textbf{J}}^t_\textbf{d}$ as, 
\begin{gather}
\renewcommand\arraystretch{1.6}
\setlength{\arraycolsep}{3pt}
    \widehat{\textbf{J}}^t_\textbf{d} = \left[\, \dotsi D_k \dotsi \,\right]\in \mathbb{R}^{M\times H}\\
    \text{where $D_k$ is the $k_\text{th}$ column of $\textbf{D}$, and $k$ needs to ensure that $d_i^k\in\textbf{d}_h$.}\notag
\end{gather}
Next, we define the weighting matrix $\widehat{\textbf{W}}$ as 
\begin{equation}
\widehat{\textbf{W}}= \text{diag}(\dotsi,w_k^1,w_k^2, \dotsi, w_H^1,w_H^2, \sigma_1, \dotsi, \sigma_M)
\end{equation}
where $k$ needs to ensure $d_i^k$ is invalid (high error), and $\sigma$ equals to $\alpha_1$ or $\alpha_2$, depending on the corresponding disparity is invalid or valid (\ie same $\alpha_1$ and $\alpha_2$ as used in \cref{eq:supp_wq_loss}). Then, similar to \cref{eq:solve_gn}, we use the Gauss-Newton method to form a linear equation,
\begin{equation}
\renewcommand\arraystretch{1.8}
\setlength{\arraycolsep}{4pt}
    \begin{bmatrix}
        \widehat{\textbf{J}}_\textbf{s}^T \widehat{\textbf{W}} \widehat{\textbf{J}}_\textbf{s} & \widehat{\textbf{J}}_\textbf{s}^T \widehat{\textbf{W}}\widehat{\textbf{J}}_\textbf{d} \\
        \widehat{\textbf{J}}_\textbf{d}^T \widehat{\textbf{W}}\widehat{\textbf{J}}_\textbf{s} & \widehat{\textbf{J}}_\textbf{d}^T \widehat{\textbf{W}}\widehat{\textbf{J}}_\textbf{d} 
    \end{bmatrix}
    \begin{bmatrix}
        \Delta \textbf{s} \\ \Delta \textbf{d} 
    \end{bmatrix}
    = - \widehat{\textbf{J}}^T\widehat{\textbf{W}}\hat{\textbf{r}}\enspace,
\end{equation}
and again use the Schur Complement \cref{eq:schur} to solve for $\Delta \textbf{s}$ and $\Delta \textbf{d}$. Note that the lower right corner block $\widehat{\textbf{J}}_\textbf{d}^T \widehat{\textbf{W}}\widehat{\textbf{J}}_\textbf{d}$ is still diagonal. Therefore, it is easy to compute its inverse.

\subsection{Interpolation of Neural Point Cloud}
Basically the strategy we use for feature interpolation in the neural point cloud is the same as in Point-SLAM~\cite{sandstrom2023point}, but for the convenience of readers, we also provide it here. In this section, we reuse the notation used in \cite{sandstrom2023point} for simplicity. Note that we redefine variables that may have been used before in this section.

As mentioned in the main paper, we use the same architecture for the occupancy decoder $\mathcal{H}$ and the color decoder $\mathcal{G}_\xi$ as~\cite{sandstrom2023point} and use their provided pretrained and fixed middle geometric decoder $\mathcal{H}$. The decoder input is the 3D point $x_i$, to which
we apply a learnable Gaussian positional encoding~\cite{tancik2020fourier} to mitigate the limited band-width of MLPs, and the associated feature. We further denote $P^g(x_i)$ and $P^c(x_i)$ as the
geometric and color features extracted at point $x_i$ respectively. For each point $p_i$ we use the corresponding per-pixel query radius $2r$, where r is computed according to Eq.\,(\textcolor{red}{2}).
Within the radius $2r$, we require to find at least two neighbors. Otherwise, the point is given zero occupancy. We use the closest eight neighbors and use inverse squared distance weighting for the geometric features, \ie
\begin{equation}
    P^g(x_i)=\sum_k \dfrac{w_k}{\sum_k w_k}f^g_k \quad \text{with} \enspace w_k = \dfrac{1}{\|p_k-x_i\|^2}
    \enspace.
\end{equation}
The color features are obtained similarly,
\begin{equation}
    P^c(x_i)=\sum_k \dfrac{w_k}{\sum_k w_k}f^c_k \quad \text{with} \enspace w_k = \dfrac{1}{\|p_k-x_i\|^2}
    \enspace.
\end{equation}
Note that we do not use the non-linear pre-processing on the extracted neighbor features as done in \cite{sandstrom2023point} since we found this to be unstable in the RGB-only setting. Next, we describe how the per-point occupancies $s_i$ and colors $t_i$ are used to render the per-pixel depth and color using volume rendering. We construct a weighting function, $\alpha_i$ as described in \cref{eq:ray_marching}. This weight represents the discretized probability that the ray terminates at point $x_i$.
\begin{equation}
    \alpha_i = s_{p_i}\prod_{j=1}^{i-1} (1-s_{p_j})\enspace.
    \label{eq:ray_marching}
\end{equation}
The rendered depth is computed as the weighted average of the depth values along each ray, and equivalently for the color according to \cref{eq:supp_rgbd-render},
\begin{align}
   D_{\text{render}} = \sum_{i=1}^N \alpha_i z_i  \enspace, \quad 
   I_{\text{render}} = \sum_{i=1}^N \alpha_i t_i  \enspace,
   \label{eq:supp_rgbd-render}
\end{align}


\section{More Experiments}
To accompany the evaluations provided in the main paper, we provide further experiments in this section. 
\subsection{Full Evaluations Data}
\begin{table}[!ht]
    \def\dashline{\noalign{\vskip 3pt} \cdashline{2-11}\noalign{\vskip 3pt}}
    \centering
    \scriptsize
    \setlength{\tabcolsep}{2.68pt}
    \resizebox{\columnwidth}{!}
    {
    \begin{tabularx}{\linewidth}{lllccccccccc}
    \toprule
     & & Metric & \texttt{R-0} & \texttt{R-1} & \texttt{R-2} & \texttt{O-0} & \texttt{O-1} & \texttt{O-2} & \texttt{O-3} & \texttt{O-4} & Avg.\\
    \midrule
    \multicolumn{2}{l}{\multirow{4}{*}{\rotatebox{0}{\makecell[l]{Reconstruction}}}}
    & Depth L1 $\downarrow$ & 2.98 & 2.52 & 3.30 & 3.29 & 2.76 & 3.72 & 4.16 & 3.11 & \textbf{3.24}\\
    & &Accuracy $\downarrow$ & 2.84 & 3.07 & 3.05 & 2.98 & 2.06 & 3.32 & 3.34 & 2.92 & \textbf{2.96}\\
    & &Completion $\downarrow$ & 4.65 & 3.55 & 3.64 & 2.39 & 3.43 & 4.54 & 4.57 & 4.78 & \textbf{3.95}\\
    & &Comp. Rat. $\uparrow$ & 81.96 & 85.78 & 84.50 & 88.82 & 85.07 & 82.09 & 80.41 & 81.04 & \textbf{83.72}\\

    \midrule
    \multirow{6}{*}{\rotatebox{0}{\makecell[l]{Render-\\ing}}}
    & \multirow{3}{*}{\rotatebox{0}{\makecell[l]{Key\\Frames}}}
      &PSNR $\uparrow$&28.49 & 30.09 & 29.98 & 35.88 & 37.15 & 28.45 & 28.54 & 29.73& \textbf{31.04} \\
    & &SSIM $\uparrow$&0.96 & 0.97 & 0.96 & 0.98 & 0.99 & 0.97 & 0.97 & 0.97& \textbf{0.97}\\
    & &LPIPS $\downarrow$&0.13 & 0.13 & 0.14 & 0.09 & 0.08 & 0.15 & 0.11 & 0.15& \textbf{0.12}\\

    \noalign{\vskip 3pt} \cdashline{2-12}\noalign{\vskip 3pt}
    & \multirow{3}{*}{\rotatebox{0}{\makecell[l]{Every\\5 Frames}}}
      &PSNR $\uparrow$&27.15 & 28.85 & 28.56 & 33.95 & 36.27 & 26.78 & 26.95 & 28.11& \textbf{29.58}\\
    & &SSIM $\uparrow$&0.95 & 0.96 & 0.95 & 0.97 & 0.99 & 0.96 & 0.96 & 0.95& \textbf{0.96}\\
    & &LPIPS $\downarrow$&0.15 & 0.12 & 0.16 & 0.11 & 0.08 & 0.17 & 0.13 & 0.17& \textbf{0.14}\\
    


\midrule
\renewcommand{\arraystretch}{2}
    \multirow{4}{*}{\rotatebox{0}{\makecell[l]{Track-\\ing}}}
    & \multirow{2}{*}{\makecell[l]{Key Frames\\Trajectory}}
      &\multirow{2}{*}{\makecell[l]{ATE\\ RMSE} $\downarrow$}
      &\multirow{2}{*}{0.31}&\multirow{2}{*}{0.52}&\multirow{2}{*}{0.21}&\multirow{2}{*}{0.26}&\multirow{2}{*}{0.29}&\multirow{2}{*}{0.41}&\multirow{2}{*}{0.46}&\multirow{2}{*}{0.44}&\multirow{2}{*}{\textbf{0.36}}
      \\ \\ \noalign{\vskip 3pt} \cdashline{2-12}\noalign{\vskip 3pt}
    & \multirow{2}{*}{\makecell[l]{Full\\Trajectory}}
      &\multirow{2}{*}{\makecell[l]{ATE\\ RMSE} $\downarrow$}
      &\multirow{2}{*}{0.31}&\multirow{2}{*}{0.37}&\multirow{2}{*}{0.20}&\multirow{2}{*}{0.29}&\multirow{2}{*}{0.28}&\multirow{2}{*}{0.45}&\multirow{2}{*}{0.45}&\multirow{2}{*}{0.44}&\multirow{2}{*}{\textbf{0.35}}
    \\ \\

\bottomrule
\end{tabularx}
    }
\caption{\textbf{Full Evaluation on Replica~\cite{straub2019replica}.} For \texttt{Keyframes} we only measure the performance on the frames which are used for mapping, which is the same strategy as existing methods. We also provide rendering results every 5 frames regardless whether they are keyframes or not, shown as \texttt{Every\;5\;Frames}. We show the ATE RMSE [cm] evaluation on the keyframes as well as on the full trajectory.} 
\label{tab:replica_full}
\vspace{-1em}
\end{table}

\begin{table}[!ht]
    \def\dashline{\noalign{\vskip 3pt} \cdashline{2-11}\noalign{\vskip 3pt}}
    \centering
    \scriptsize
    \setlength{\tabcolsep}{5.02pt}
    \begin{tabularx}{\linewidth}{lllcccccc}
    \toprule
     & & Metric & \texttt{f1/desk} & \texttt{f1/desk2} & \texttt{f1/room} & \texttt{f2/xyz} & \texttt{f3/office} & Avg.\\

    \midrule
    \multirow{6}{*}{\rotatebox{0}{\makecell[l]{Render-\\ing}}}
    & \multirow{3}{*}{\rotatebox{0}{\makecell[l]{Key\\Frames}}}
      &PSNR $\uparrow$&20.26 & 19.09 & 18.78 & 25.62 & 21.21 & 
      \textbf{20.99}\\
    & &SSIM $\uparrow$&0.87 & 0.82 & 0.79 & 0.96 & 0.84 & \textbf{0.85}\\
    & &LPIPS $\downarrow$&0.31 & 0.38 & 0.38 & 0.09 & 0.32 & \textbf{0.30}\\
    \noalign{\vskip 3pt} \cdashline{2-9}\noalign{\vskip 3pt}
    & \multirow{3}{*}{\rotatebox{0}{\makecell[l]{Every\\5 Frames}}}
      &PSNR $\uparrow$&17.16 & 15.93 & 16.17 & 21.72 & 17.25 & \textbf{17.65}\\
    & &SSIM $\uparrow$&0.79 & 0.72 & 0.72 & 0.92 & 0.73 & \textbf{0.77}\\
    & &LPIPS $\downarrow$&0.40 & 0.44 & 0.42 & 0.15 & 0.44 & \textbf{0.37}\\

\midrule
\renewcommand{\arraystretch}{2}
    \multirow{4}{*}{\rotatebox{0}{\makecell[l]{Track-\\ing}}}
    & \multirow{2}{*}{\makecell[l]{Key Frames\\Trajectory}}
      &\multirow{2}{*}{\makecell[l]{ATE\\ RMSE} $\downarrow$}
      &\multirow{2}{*}{1.92}&\multirow{2}{*}{3.05}&\multirow{2}{*}{4.43}&\multirow{2}{*}{0.23}&\multirow{2}{*}{1.41}&\multirow{2}{*}{\textbf{2.21}}
      \\ \\ \noalign{\vskip 3pt} \cdashline{2-9}\noalign{\vskip 3pt}
    & \multirow{2}{*}{\makecell[l]{Full\\Trajectory}}
      &\multirow{2}{*}{\makecell[l]{ATE\\ RMSE} $\downarrow$}
      &\multirow{2}{*}{1.65}&\multirow{2}{*}{2.79}&\multirow{2}{*}{4.16}&\multirow{2}{*}{0.22}&\multirow{2}{*}{1.44}&\multirow{2}{*}{\textbf{2.05}}
    \\ \\
\bottomrule
\end{tabularx}
\caption{\textbf{Full Evaluation on TUM-RGBD~\cite{Sturm2012ASystems}.}} 
\label{tab:tum_full}
    \vspace{-1em}
\end{table}

\begin{table}[!ht]
    \def\dashline{\noalign{\vskip 3pt} \cdashline{2-11}\noalign{\vskip 3pt}}
    \centering
    \scriptsize
    \setlength{\tabcolsep}{3.42pt}
    \begin{tabularx}{\linewidth}{lllccccccccc}
    \toprule
     & & Metric & \texttt{0000} & \texttt{0054} & \texttt{0059} & \texttt{0106} & \texttt{0169} & \texttt{0181} & \texttt{0207} & \texttt{0233} & Avg.\\

    \midrule
    \multirow{6}{*}{\rotatebox{0}{\makecell[l]{Render-\\ing}}}
    & \multirow{3}{*}{\rotatebox{0}{\makecell[l]{Key\\Frames}}}
    &PSNR$\uparrow$&23.42&25.68&20.66&20.41&25.23&21.28&23.68&22.29&\textbf{22.83}\\
    & &SSIM $\uparrow$&0.87&0.87&0.83&0.84&0.91&0.76&0.85&0.84&\textbf{0.84}\\
    & &LPIPS $\downarrow$&0.26&0.30&0.31&0.31&0.21&0.44&0.29&0.30&\textbf{0.30}\\
    \noalign{\vskip 3pt} \cdashline{2-12}\noalign{\vskip 3pt}
    & \multirow{3}{*}{\rotatebox{0}{\makecell[l]{Every\\5 Frames}}}
    &PSNR$\uparrow$&21.80&23.75&18.66&17.54&22.85&19.14&21.52&20.36&\textbf{20.70}\\
    & &SSIM $\uparrow$&0.83&0.83&0.77&0.77&0.88&0.71&0.81&0.78&\textbf{0.80}\\
    & &LPIPS $\downarrow$&0.30&0.34&0.36&0.42&0.25&0.47&0.32&0.36&\textbf{0.35}\\

\midrule
\renewcommand{\arraystretch}{2}
    \multirow{4}{*}{\rotatebox{0}{\makecell[l]{Track-\\ing}}}
    & \multirow{2}{*}{\makecell[l]{Key Frames\\Trajectory}}
      &\multirow{2}{*}{\makecell[l]{ATE\\ RMSE} $\downarrow$}
      &\multirow{2}{*}{5.66}&\multirow{2}{*}{9.17}&\multirow{2}{*}{9.48}&\multirow{2}{*}{7.03}&\multirow{2}{*}{8.72}&\multirow{2}{*}{8.42}&\multirow{2}{*}{7.47}&\multirow{2}{*}{4.97}&\multirow{2}{*}{\textbf{7.61}}
      \\ \\ \noalign{\vskip 3pt} \cdashline{2-12}\noalign{\vskip 3pt}
    & \multirow{2}{*}{\makecell[l]{Full\\Trajectory}}
      &\multirow{2}{*}{\makecell[l]{ATE\\ RMSE} $\downarrow$}
      &\multirow{2}{*}{5.57}&\multirow{2}{*}{9.50}&\multirow{2}{*}{9.11}&\multirow{2}{*}{7.09}&\multirow{2}{*}{8.26}&\multirow{2}{*}{8.39}&\multirow{2}{*}{7.53}&\multirow{2}{*}{5.17}&\multirow{2}{*}{\textbf{7.58}}
    \\ \\

\bottomrule
\end{tabularx}
\caption{\textbf{Full Evaluation on ScanNet~\cite{Dai2017ScanNet}.}} 
\label{tab:scannet_full}
    \vspace{-2em}
\end{table}

\noindent
In \cref{tab:replica_full}, \cref{tab:tum_full} and \cref{tab:scannet_full}, we provide the full per scene results on all commonly reported metrics on Replica~\cite{straub2019replica}, TUM-RGBD~\cite{Sturm2012ASystems} and ScanNet~\cite{Dai2017ScanNet}. 

The reconstruction results are only measured on Replica since the other two datasets are real world datasets which lack quality ground truth meshes. 

For the rendering metrics, we not only report the results on the keyframes (the frames used for mapping), but also the every-5-frames results. The every-5-frames rendering performance is very close compared to the keyframe rendering results, which suggests that our method can generalize to unseen views.

We also show the trajectory accuracy measurement of both keyframes and all the full trajectory, which is obtained by first linear interpolation between keyframes and using optical flow to refine. The accuracy of these two trajectories are similar. In the main paper, the data we report is always measured on the full trajectory.

\subsection{Influence of Monocular Depth Prior}

\begin{table}[t]
    \def\dashline{\noalign{\vskip 3pt} \cdashline{2-11}\noalign{\vskip 3pt}}
    \centering
    \scriptsize
    \setlength{\tabcolsep}{4.18pt}
    \resizebox{\columnwidth}{!}
    {
    \begin{tabularx}{\linewidth}{lllccccccccc}
    \toprule
     & & Metric & \texttt{R-0} & \texttt{R-1} & \texttt{R-2} & \texttt{O-0} & \texttt{O-1} & \texttt{O-2} & \texttt{O-3} & \texttt{O-4} & Avg.\\
    \midrule
    \multicolumn{2}{c}{\multirow{4}{*}{\rotatebox{0}{\makecell[l]{Recon-\\struction}}}}
    & Depth L1 $\downarrow$ & 2.16 & 1.41 & 1.85 & 1.64 & 1.77 & 3.23 & 5.71 & 1.93 & \textbf{2.46}\\
    & &Accuracy $\downarrow$ & 2.02 & 1.31 & 1.43 & 1.22 & 0.95 & 3.47 & 4.99 & 1.77 & \textbf{2.15}\\
    & &Completion $\downarrow$ & 3.26 & 2.71 & 2.59 & 1.56 & 2.19 & 3.80 & 3.23 & 3.52 & \textbf{2.86}\\
    & &Comp. Rat. $\uparrow$ & 88.27 & 89.98 & 89.65 & 94.06 & 91.11 & 85.17 & 85.19 & 85.65 & \textbf{88.64}\\

    \midrule
    \multicolumn{2}{c}{\multirow{3}{*}{\rotatebox{0}{\makecell[l]{Rendering}}}}
      &PSNR $\uparrow$&29.98 & 31.98 & 32.86 & 37.29 & 38.03 & 30.34 & 29.40 & 33.37& \textbf{32.91} \\
    & &SSIM $\uparrow$&0.97 & 0.98 & 0.98 & 0.99 & 0.99 & 0.98 & 0.98 & 0.98& \textbf{0.98}\\
    & &LPIPS $\downarrow$&0.10 & 0.11 & 0.09 & 0.08 & 0.06 & 0.13 & 0.10 & 0.12& \textbf{0.10}\\

\midrule
\renewcommand{\arraystretch}{2}
    \multirow{2}{*}{\rotatebox{0}{\makecell[c]{Tracking}}}

    &
      &\multirow{2}{*}{\makecell[l]{ATE\\ RMSE} $\downarrow$}
      &\multirow{2}{*}{0.38}&\multirow{2}{*}{0.36}&\multirow{2}{*}{0.24}&\multirow{2}{*}{0.24}&\multirow{2}{*}{0.45}&\multirow{2}{*}{0.37}&\multirow{2}{*}{0.42}&\multirow{2}{*}{0.44}&\multirow{2}{*}{\textbf{0.36}}
    \\ \\

\bottomrule
\end{tabularx}
    }
\caption{\textbf{Full Evaluations on Replica~\cite{straub2019replica} with ground truth depth.} Both reconstruction and rendering results improve significantly with the ground truth depth, suggesting that our method is bounded by the quality of current day monocular depth estimation. Since we do not require any extra training or fine-tuning of the monocular depth estimator, it is easy to plug in a better estimator once available. Tracking performance does not change much.} 
\label{tab:replica_full_rgbd}
    \vspace{-1em}
\end{table}

While we show that the monocular depth prior improves the geometric estimation capability of our framework, it may still be erroneous. To better understand the accuracy of monocular depth prior, we replace the monocular depth prior with the ground truth sensor depth instead. This experiment acts as the upper bound of our method if the monocular depth prior is perfect. The experiments are done on Replica~\cite{straub2019replica} and are shown in \cref{tab:replica_full_rgbd}. Compared with the standard setting with the monocular prior, the ground truth depth setting gives improvements on both reconstruction and rendering quality, which reveals that our method still has potential to achieve better mapping results once a better monocular depth prior is available. Since our method does not require further training or fine-tuning for the monocular depth prior, it is quite easy to just replace the current off-the-shelf monocular depth estimator with a better one.

\subsection{Scale and Shift in Depth Space or Disparity Space}
\begin{table}[!ht]
    \centering
    \scriptsize
    \setlength{\tabcolsep}{8.6pt}
    \begin{tabularx}{\linewidth}{lccccccc}
    \toprule
      &\makecell[c]{PSNR\\$[dB]\uparrow$} & \makecell[c]{Depth L1\\$[cm]\downarrow$} &\makecell[c]{Accuracy\\$[cm]\downarrow$} &\makecell[c]{Completion\\$[cm]\downarrow$}& \makecell[c]{Completion Ratio\\$[\%]\uparrow$}\\
    \midrule
    Disparity Space & 27.96 & 2.53 & 3.15 & 3.61 & 85.61\\  
    Depth Space & 30.09 & 2.52 & 3.07 & 3.55 & 85.78\\  
           
    \bottomrule
    \end{tabularx}
    \caption{
    \textbf{Comparison of Scale and Shift Alignment in Different Spaces on Replica~\cite{straub2019replica} \texttt{room1}.}  There is marginal performance difference between disparity and depth alignment. Using depth alignment is marginally better in all metrics.
    }
    \label{tab:disp_vs_depth}
    \vspace{-2.5em}
\end{table}
\noindent
In our paper, we align the scale and shift of the monocular depth prior in depth space with the valid estimated depth, as shown in Eq.\,(\textcolor{red}{8}) in the main paper. However, there are some other works~\cite{ranftl2020towards} that suggest that the alignment should be done in disparity space. We therefore compare the difference between using disparity space or depth space alignment. The results are shown in \cref{tab:disp_vs_depth}. The results show that in general the space choice does not influence the results much, especially for geometry, but depth space still performs a bit better. Thus, we use depth space to do the scale and shift alignment in our proposed method.

\begin{figure}[!ht]
\vspace{0em}
\centering
{
\setlength{\tabcolsep}{1pt}
\renewcommand{\arraystretch}{1}
\newcommand{\sz}{0.3}
\newcommand{\subsz}{0.2}
\begin{tabular}{ccccc}
\multirow{2}{*}[2em]{\rotatebox{90}{\texttt{scene 0054}}}
&
\includegraphics[width=\sz\linewidth]{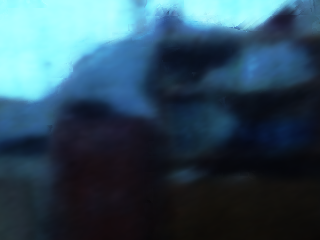} &
\includegraphics[width=\sz\linewidth]{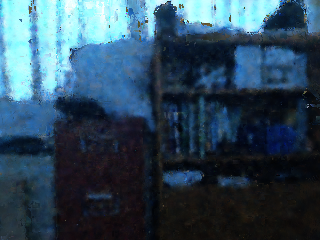} &
\includegraphics[width=\sz\linewidth]{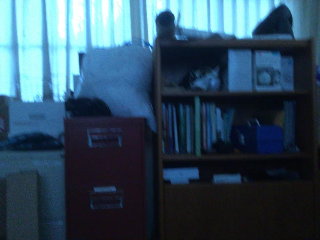}
\\
&
\includegraphics[width=\sz\linewidth]{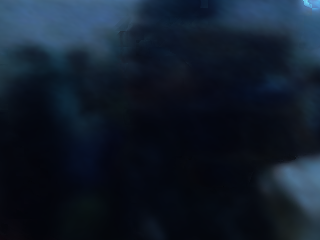} &
\includegraphics[width=\sz\linewidth]{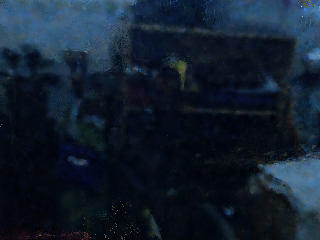} &
\includegraphics[width=\sz\linewidth]{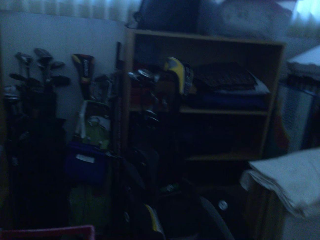}
\\
\multirow{2}{*}[2em]{\rotatebox{90}{\texttt{scene 0169}}}&
\includegraphics[width=\sz\linewidth]{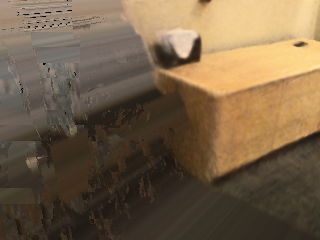} &
\includegraphics[width=\sz\linewidth]{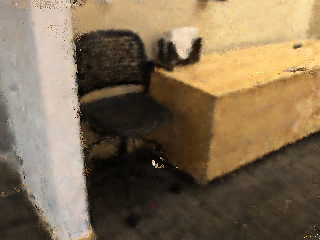} &
\includegraphics[width=\sz\linewidth]{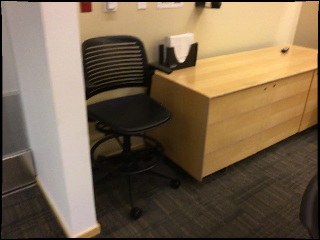}
\\
&
\includegraphics[width=\sz\linewidth]{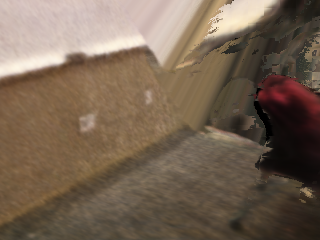} &
\includegraphics[width=\sz\linewidth]{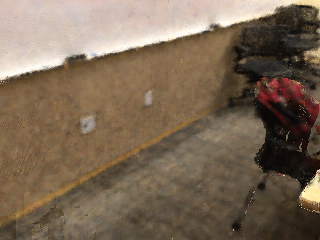} &
\includegraphics[width=\sz\linewidth]{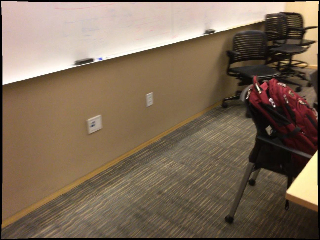}
\\
\multirow{2}{*}[2em]{\rotatebox{90}{\texttt{scene 0233}}}&
\includegraphics[width=\sz\linewidth]{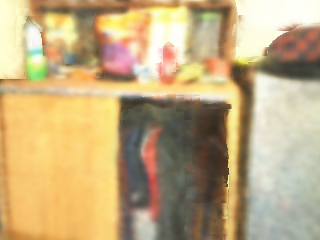} &
\includegraphics[width=\sz\linewidth]{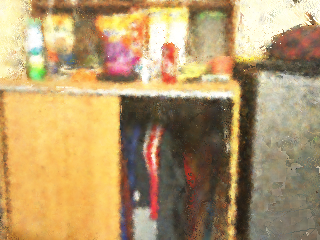} &
\includegraphics[width=\sz\linewidth]{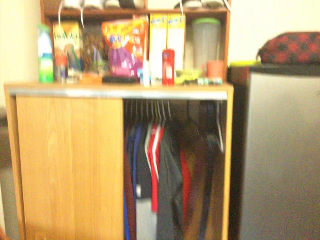}
\\
&
\includegraphics[width=\sz\linewidth]{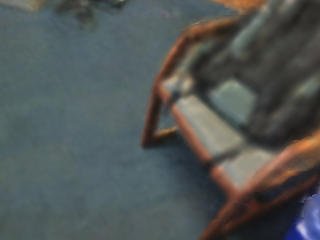} &
\includegraphics[width=\sz\linewidth]{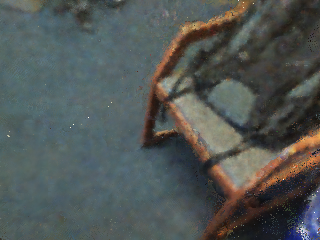} &
\includegraphics[width=\sz\linewidth]{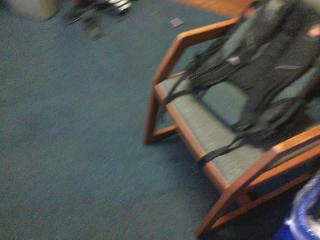}
\\

\Large
&GO-SLAM~\cite{zhang2023go}  & \textbf{\ours (ours)} & Ground Truth\\
\end{tabular}
}
\caption{\textbf{Additional Rendering Results on ScanNet~\cite{Dai2017ScanNet}.} Thanks to the deformable neural point cloud, our method can maintain a globally consistent map and render high-frequency details with high fidelity.}
\label{fig:render_scannet_addition}
\vspace{0em}
\end{figure}

\subsection{Additional Qualitative Renderings}
In \cref{fig:render_scannet_addition} we show additional renderings from the ScanNet dataset where our method is compared to GO-SLAM~\cite{zhang2023go}.

\clearpage

%
%
{\small
\bibliographystyle{splncs04}
\bibliography{main}
}
\end{document}